\documentclass[11pt]{article}
\usepackage[a4paper,margin=1in]{geometry}
\usepackage[T1]{fontenc}
\usepackage{lmodern}
\usepackage{amsmath,amsthm}
\usepackage{microtype}
\usepackage{graphicx}
\usepackage{booktabs}
\usepackage{enumitem}
\usepackage{flafter}
\usepackage{placeins}
\usepackage[round,authoryear]{natbib}
\usepackage{xcolor}
\usepackage{mymacro}
\usepackage[colorlinks=true,linkcolor=blue!55!black,citecolor=green!35!black,urlcolor=blue!55!black]{hyperref}
\usepackage[nameinlink,capitalize,noabbrev]{cleveref}
\hypersetup{pdftitle={Instance-Optimal Moment Estimation in the Add-Remove Model of Differential Privacy},pdfauthor={Shokichi Takakura},pdfsubject={Sharp local minimax estimation of empirical means, variances, and covariances}}
\newcommand{\risk}{\mathcal{R}}

\newcommand{\Ball}[2]{B_{#1}(#2)}
\newcommand{\ve}{v_{\varepsilon}}

\graphicspath{{figures/}}
\setlist[itemize]{topsep=4pt,itemsep=3pt,leftmargin=*}
\title{Instance-Optimal Moment Estimation\\in the Add-Remove Model of Differential Privacy}
\author{Shokichi Takakura\\LY Corporation}
\date{}
\begin{document}
\maketitle
\begin{abstract}
    In this paper, we study the problem of estimating empirical means, variances, and covariances
    under pure differential privacy in the add-remove model.
    While existing mechanisms achieve optimal worst-case error,
    they may not be optimal for individual datasets.
    To measure the performance of mechanisms on individual datasets, we introduce instance-dependent error metrics
    and prove sharp lower bounds on them.
    Then, we design mechanisms that achieve these lower bounds
    and demonstrate that they achieve asymptotically better utility compared to existing methods.
\end{abstract}
\section{Introduction}
Estimating the mean, variance, and covariance of a dataset is a fundamental task in statistical analysis.
They are widely used to summarize and understand the characteristics of the data.
However, releasing statistics directly may leak sensitive information about individuals in the dataset,
for instance, through difference attacks~\citep{dalenius1977towards}.

Differential Privacy (DP)~\citep{dwork2006calibrating,dwork2014algorithmic} provides a formal framework to quantify and control the privacy leakage
and is used in practical systems for privacy-preserving data analysis~\citep{mcsherry2009privacy,wilson2019differentially,rogers2020linkedin,amin2022plume}.
It ensures that the output of a computation does not significantly change for two neighboring datasets that differ by a single record.
In the definition of neighboring datasets, there are two common models: \textit{the swap model} and \textit{the add-remove model}.
In the swap model, neighboring datasets differ by swapping a single record with another,
and in the add-remove model, neighboring datasets differ by the insertion or deletion of a single record.
Since the add-remove model protects not only the record values but also the participation of individuals in the dataset,
it is particularly suitable for scenarios where the presence or absence of a single record is sensitive information~\citep{dwork2014algorithmic}.

Despite its advantages, the add-remove model poses additional challenges for statistical estimation,
since the number of records in the dataset is itself sensitive information and must be protected.
Recently,~\citet{kulesza2024mean} studied the problem of mean estimation in the add-remove model of differential privacy and proposed the transformed noise addition,
which is shown to achieve the optimal worst-case asymptotic error for mean estimation in the add-remove model.
Furthermore,~\citet{takakura2025optimal} extended it to the problem of variance and covariance estimation
and developed B\'ezier mechanisms, which achieve optimal worst-case asymptotic error for variance and covariance estimation in the high-privacy regime.

While worst-case optimality provides valuable insight into algorithm performance,
it may not capture the performance of mechanisms on individual datasets.
For example, consider a mechanism with error $e_1$ on every dataset and another with error $e_1$ on some datasets and $e_2<e_1$ on others.
Both have worst-case error $e_1$, so worst-case analysis cannot distinguish them despite the second mechanism's improved accuracy on some datasets.
This limitation motivates instance-dependent optimality criteria that account for such improvements~\citep{asi2020near,huang2021instance,dick2023subset,durfee2025unifying}.
Considering the importance of instance-optimality, the fundamental question naturally arises:
\textit{What is the best achievable accuracy for mean, variance, and covariance estimation for individual datasets in the add-remove model? Is there any algorithm that can achieve it uniformly across all datasets?}

In this work, we show that there exist instance-optimal mechanisms for mean, variance, and covariance estimation
which achieve uniformly better utility than existing worst-case optimal mechanisms such as the transformed noise addition and B\'ezier mechanisms for sufficiently large datasets.
Specifically, we define local minimax error to characterize instance-dependent performance
and prove a sharp lower bound on it.
Then, we develop a novel mechanism, which achieves this lower bound, thereby establishing instance-optimality.
Technically, we develop a self-calibration technique which leverages preliminary estimates to adjust the noise scale
and adaptive budget allocation to distribute the privacy budget between the preliminary and refinement phases adaptively to the dataset size.

Our contributions can be summarized as follows:
\begin{itemize}
    \item We define local minimax error to characterize instance-optimal performance and establish sharp asymptotic local lower bounds for all private mechanisms.
    \item We construct two-phase estimators by developing two technical components: self-calibration and adaptive budget allocation.
          We prove that the proposed estimators achieve the local lower bounds, thereby establishing their instance-optimality.
    \item We compare the proposed estimators with transformed noise addition~\citep{kulesza2024mean} and B\'ezier mechanism~\citep{takakura2025optimal}.
          We show that the proposed estimators asymptotically achieve lower error than these existing mechanisms, especially on datasets whose mean is close to the center of the data range.
\end{itemize}

\subsection{Related Work}

\paragraph{Instance optimality and data-dependent accuracy.}
Beyond worst-case analysis, several works have explored instance-dependent error analysis~\citep{asi2020near,huang2021instance,dick2023subset,durfee2025unifying}.
In addition, a line of work~\citep{nissim2007smooth,bun2019average} considers the notion of smooth sensitivity for dataset-dependent noise calibration.
However, most existing works focus on the swap model of differential privacy
or establish optimality only up to constant or logarithmic factors.
On the other hand, we establish sharp asymptotic local minimax optimality in the add-remove model,
including the exact leading constants.

\paragraph{Budget selection.}
Gradual release~\citep{koufogiannis2015gradual} constructs correlated Laplace noise with increasing privacy budget, with the overall privacy cost bounded by the largest budget.
\citet{ligett2017accuracy} utilize gradual release to adaptively select privacy budgets by evaluating intermediate outputs and stopping when a desired accuracy is achieved.
Our adaptive allocation strategy also builds on gradual release techniques.
However, it does not require specifying a minimum privacy budget in advance,
nor does it spend additional privacy budget for the stopping criterion.

\section{Problem Settings}\label{sec:problem-settings}
In this paper, we assume that the data are bounded within $[-R,R]$ for some fixed $R>0$ and
consider the problem of estimating the empirical mean, variance, and covariance defined as
\begin{align*}
    \mu=\frac1n\sum_i x_i,\qquad
    v=\frac1n\sum_i(x_i-\mu)^2,\qquad
    c=\frac1n\sum_i(x_i-\mu_x)(y_i-\mu_y),
\end{align*}
where $x \in [-R, R]^n$ and $y \in [-R, R]^n$.
These statistics satisfy $\mu\in[-R,R]$, $v\in[0,R^2]$, and $c\in[-R^2,R^2]$.

For $f\in\{\mu,v,c\}$, let $\cD_f$ contain all finite multisets over its data domain.
Two datasets $D,E\in\cD_f$ are defined to be neighboring, denoted by $D\sim E$, if one is obtained from the other by adding or removing one record.

\begin{definition}[Differential privacy~\citep{dwork2014algorithmic}]\label[definition]{def:differential-privacy}
    Let $\varepsilon\ge0$ and let $A:\cD_f\to\cY$ be a randomized mechanism with measurable output space $\cY$.
    The mechanism satisfies $\varepsilon$-differential privacy if, for every pair $D\sim E$ and every measurable set $B\subseteq\cY$,
    \begin{align*}
        \Prob{A(D)\in B}\le e^\varepsilon\Prob{A(E)\in B}.
    \end{align*}
\end{definition}

Let $\cA_\varepsilon^f$ denote the class of all real-valued $\varepsilon$-DP mechanisms on $\cD_f$.
For notational simplicity, we omit the index $f$ when the statistic is clear from context.
Then, define $L_f(A,D)=\Expec{(A(D)-f(D))^2}$ as the mean squared error of a mechanism $A$ on a dataset $D$.

Existing works~\citep{kulesza2024mean,takakura2025optimal} have studied the global minimax risk defined as follows:
\begin{definition}[Normalized global minimax risk]\label[definition]{def:global-minimax-risk}
    For an integer $n\ge1$, the normalized global minimax risk is defined as
    \begin{align*}
        \risk^f_n=\inf_{A\in\cA_\varepsilon^f}\sup_{\substack{D\in\cD_f \\|D|\ge n}}|D|^2L_f(A,D).
    \end{align*}
\end{definition}
This criterion measures the worst-case error over all datasets of size at least $n$.
However, the global minimax risk may be overly pessimistic, as it considers the worst-case error over all datasets of size at least $n$,
including those far from the dataset of interest.

To study the performance of the mechanism for a specific dataset, we introduce the notion of normalized local minimax risk.
\begin{definition}[Normalized local risk]\label[definition]{def:local-minimax-risk}
    Let $\Ball{k}{D}$ be the set of datasets obtained from $D$ by at most $k$ insertions and deletions.
    For $|D|=n$ and an integer $0\le k<n$, the normalized minimax local risk is defined as
    \begin{align}
        \risk^f_{n,k}(D)=\inf_{A\in\cA_\varepsilon^f}\sup_{E\in\Ball{k}{D}}n^2L_f(A,E).\label{eq:local-minimax-risk}
    \end{align}
\end{definition}
\label{sec:local-risk-interpretation}
If $k=0$, a mechanism which outputs a constant value $f(D)$ achieves zero risk.
Therefore, we instead consider a local neighborhood of the dataset,
allowing for a more refined assessment of the mechanism's performance near the dataset of interest.

\section{Preliminaries}\label{sec:preliminaries}
In this section, we first introduce Laplace and staircase mechanisms, which are core components in the existing methods and proposed methods.
Then, we review the existing mechanisms for mean, variance, and covariance estimation under differential privacy in the add-remove model
and derive their local error bounds.

\subsection{Laplace and Staircase Mechanisms}\label{sec:laplace-and-staircase-mechanisms}
For a query $q:\cD_f\to\R^d$, its $\ell_1$-sensitivity is defined as
\begin{align*}
    \sup_{D\sim E}\|q(D)-q(E)\|_1.
\end{align*}

The Laplace mechanism adds noise drawn from the Laplace distribution $\Lap(b)$ with density $(2b)^{-1}\exp(-|z|/b)$.
The Laplace mechanism satisfies $\varepsilon$-differential privacy by adding noise $\Lap(\Delta/\varepsilon)$ to a query with sensitivity at most $\Delta$.

To improve utility,~\citet{geng2015optimal} introduced the staircase mechanism, which adds noise with a piecewise-constant density.
For $\eta>0$, define
\begin{align*}
    q_\eta=e^{-\eta},\qquad
    h_\eta=\left(\frac{q_\eta(1+q_\eta)}{2}\right)^{1/3},\qquad
    \gamma_\eta=\frac{h_\eta-q_\eta}{1-q_\eta},\qquad
    a_\eta=\frac{1-q_\eta}{2h_\eta}.
\end{align*}
Then, the symmetric density of staircase noise $T_\eta$ is defined as
\begin{align}
    f_\eta(z)=
    \begin{cases}
        a_\eta q_\eta^j,     & j\le|z|<j+\gamma_\eta,   \\
        a_\eta q_\eta^{j+1}, & j+\gamma_\eta\le|z|<j+1.
    \end{cases}\label{eq:staircase-density}
\end{align}
The variance of staircase noise is given by
\begin{align}
    v_\eta=\Expec{T_\eta^2}
    =\frac{2^{-2/3}q_\eta^{2/3}(1+q_\eta)^{2/3}+q_\eta}{(1-q_\eta)^2}.\label{eq:staircase-variance}
\end{align}
The staircase mechanism satisfies $\eta$-DP by adding independent noise $\Delta T_\eta$ to a scalar query with sensitivity at most $\Delta$.

\subsection{Transformed Noise Addition}\label{sec:transformed-noise-addition}
For mean estimation in the add-remove model,~\citet{kulesza2024mean} introduced transformed noise addition.
First, it transforms the count $n=|D|$ and sum $s(D)=\sum_{x\in D}x$ into
\begin{align*}
    b=\left(\frac{n+s(D)/R}{2},\frac{n-s(D)/R}{2}\right).
\end{align*}
Adding independent noises $Z_1,Z_2\sim\Lap(1/\varepsilon)$ yields the noisy estimates $\hat b=b+(Z_1,Z_2)$.
Since the $\ell_1$-sensitivity of $b$ is $1$, this release satisfies $\varepsilon$-DP.
Inverting the transformation gives estimates of $n$ and $s(D)$:
\begin{align*}
    \hat n        & = \hat b_1 + \hat b_2 = n + Z_1 + Z_2,          \\
    \widehat s(D) & = R(\hat b_1 - \hat b_2) = s(D) + R(Z_1 - Z_2).
\end{align*}
Finally, the estimator of $\mu$ is defined as
\begin{align}
    \hat \mu_{\mathrm{T}}=\Pi_\mu \ab(\frac{\hat s}{\hat n}) = \Pi_\mu\left(\frac{s(D)+R(Z_1-Z_2)}{n+Z_1+Z_2}\right),\label{eq:transformed-noise-mean-estimator}
\end{align}
for $\hat n \neq 0$, where $\Pi_\mu$ denotes projection onto $[-R,R]$ and when $\hat n = 0$, we define $\hat\mu_{\mathrm{T}}=0$.

For fixed $\varepsilon>0$, its MSE on a dataset $D$ with mean $\mu=\mu(D)$ satisfies
\begin{align*}
    L_\mu(\hat \mu_{\mathrm{T}},D)\le\frac{4(R^2+\mu^2)}{\varepsilon^2n^2}+o(n^{-2}).
\end{align*}
See~\cref{app:transformed-laplace-error-analysis} for the derivation.

Replace the independent Laplace noises in Eq.~\eqref{eq:transformed-noise-mean-estimator} with $(Z_1,Z_2)\sim\operatorname{Hourglass}(\varepsilon)$, as defined in~\cref{def:hourglass-noise}.
The resulting estimator $\hat \mu_{\mathrm{H}}$ satisfies $\varepsilon$-DP~\citep{kulesza2024mean}.
Its MSE satisfies
\begin{align}
    L_\mu(\hat \mu_{\mathrm{H}},D)\le\frac{C_{\mathrm{H}}(\mu)}{n^2}+o(n^{-2}),\qquad
    C_{\mathrm{H}}(\mu)=2(R^2+\mu^2)\ve.\label{eq:hourglass-coefficient}
\end{align}
Since $\ve<2/\varepsilon^2$, hourglass noise reduces the asymptotic error compared to Laplace noise.
See~\cref{app:transformed-hourglass-error-analysis} for the derivation.

Since $C_{\mathrm{H}}(\mu)\le 4R^2\ve$ for all $\mu\in[-R,R]$,
hourglass mechanism achieves the global minimax rate as shown in~\citet{kulesza2024mean}:
\begin{align*}
    \lim_{n\to\infty}\sup_{\substack{D\in\cD_\mu \\|D|\ge n}}|D|^2L_\mu(\hat \mu_{\mathrm{H}},D)
    =\lim_{n\to\infty}\risk_n^\mu=4R^2\ve.
\end{align*}

\subsection{B\'ezier Mechanisms}\label{sec:bezier}
For variance and covariance estimation,~\citet{takakura2025optimal} have developed the B\'ezier mechanisms
by extending transformed noise addition to higher-order moments.
It first defines the B\'ezier basis vectors for variance and covariance as
\begin{align*}
    b_v(x)   & =\frac1{4R^2}\big((R-x)^2,2(R^2-x^2),(R+x)^2\big),                  \\
    b_c(x,y) & =\frac1{4R^2}\big((R-x)(R-y),(R-x)(R+y),(R+x)(R-y),(R+x)(R+y)\big).
\end{align*}
Then, it applies the Laplace mechanism to obtain $\hat b_f=b_f+Z$, where $b_f=\sum_{z\in D}b_f(z)$ and the coordinates of $Z$ are independent $\Lap(1/\varepsilon)$ noises.
Since the $\ell_1$-sensitivity of $b_f$ is $1$, $\hat b_f$ satisfies $\varepsilon$-DP.
For variance, the estimator of $v$ is defined as
\begin{align*}
    \widehat v_{\mathrm{B}}=\Pi_v\left(4R^2\left[\frac{\hat b_{v,3}}{\widehat n_v}
        -\left(\frac{\hat b_{v,2}/2+\hat b_{v,3}}{\widehat n_v}\right)^2\right]\right),
\end{align*}
where $\widehat n_v=\sum_{j=1}^3\hat b_{v,j}$
and $\Pi_v$ denotes projection onto $[0,R^2]$.

For covariance, the estimator of $c$ is defined as
\begin{align*}
    \widehat c_{\mathrm{B}}=\Pi_c\left(4R^2\left[\frac{\hat b_{c,4}}{\widehat n_c}
        -\frac{(\hat b_{c,3}+\hat b_{c,4})(\hat b_{c,2}+\hat b_{c,4})}{\widehat n_c^2}\right]\right),
\end{align*}
where $\widehat n_c=\sum_{j=1}^4\hat b_{c,j}$
and $\Pi_c$ denotes projection onto $[-R^2,R^2]$.

For $f\in\{v,c\}$, write $\widehat f_{\mathrm B}$ for the corresponding B\'ezier estimator above.
For fixed $\varepsilon>0$, its MSE satisfies
\begin{align}
    \begin{aligned}
        L_v(\hat v_{\mathrm{B}}, D) & \le\frac{C_{\mathrm{B},v}(D)}{n^2}+o(n^{-2}), \\
        L_c(\hat c_{\mathrm{B}}, D) & \le\frac{C_{\mathrm{B},c}(D)}{n^2}+o(n^{-2}),
    \end{aligned}\label{eq:bezier-mse}
\end{align}
where the coefficients are defined by
\begin{align}
    \begin{aligned}
        C_{\mathrm{B},c} & =\frac8{\varepsilon^2}\big[R^4+R^2(\mu_x^2+\mu_y^2)+(\mu_x\mu_y-c)^2\big], \\
        C_{\mathrm{B},v} & =\frac2{\varepsilon^2}\big[3(\mu^2-v)^2+2R^2(5\mu^2-v)+3R^4\big].
    \end{aligned}\label{eq:bezier-coefficients}
\end{align}
See~\cref{app:bezier-variance-error-analysis,app:bezier-covariance-error-analysis} for the derivations.

In the worst case, $C_{\mathrm{B},v}(D)=C_{\mathrm{B},c}(D)=32R^4/\varepsilon^2$.
Then, both estimators achieve the asymptotic global minimax error in the high-privacy regime, as shown by~\citet[Theorems 3.1, 5.8, and 5.9]{takakura2025optimal}.
That is, for $f\in\{v,c\}$,
\begin{align*}
    \lim_{\varepsilon\to0}\varepsilon^2\lim_{n\to\infty}
    \sup_{\substack{D\in\cD_f \\|D|\ge n}}|D|^2L_f(\widehat f_{\mathrm B},D)
    =\lim_{\varepsilon\to0}\varepsilon^2\lim_{n\to\infty}\risk_n^f
    =32R^4.
\end{align*}

\section{Sharp Local Lower Bounds}\label{sec:sharp-local-lower-bounds}
Transformed noise addition and B\'ezier mechanisms are shown to achieve the asymptotic global minimax risk.
However, as discussed in the previous section, global minimax optimality does not necessarily imply instance-wise optimality.
In this section, we derive lower bounds for the local minimax risk of the mean, variance, and covariance,
which provide fundamental limits on the accuracy of any private estimators.

First, we define the following quantities related to the effect of a single record on the target statistic.
\begin{align}
    \begin{aligned}
        \psi_{\mu,D}(x) & =x-\mu,                & \Delta_\mu & =R+|\mu|,
        \\
        \psi_{v,D}(x)   & =(x-\mu)^2-v,          & \Delta_v   & =\max\{v,(R+|\mu|)^2-v\},
        \\
        \psi_{c,D}(x,y) & =(x-\mu_x)(y-\mu_y)-c, & \Delta_c   & =\max_{x, y\in\{-R,R\}}|(x-\mu_x)(y-\mu_y)-c|.
    \end{aligned}\label{eq:moment-influence-sensitivity-coefficients}
\end{align}
Similar notion can be found in~\citet{chaudhuri2012convergence} but their focus is on the swap model.

Using the above definitions, we can show the following lower bound on the local minimax risk.
\begin{theorem}[Lower bound on local minimax risk]\label{thm:local-minimax-lower-bounds}
    Consider a sequence of datasets $D_n$ of size $n = 1, 2, \dots$
    and define $\mu_n, v_n$ and $c_n$ as the empirical mean, variance, and covariance of $D_n$, respectively.
    Suppose that $\mu_n\to\mu \in (-R, R)$ for the mean, $(\mu_n,v_n)\to(\mu,v)$ for variance, or $(\mu_{x,n},\mu_{y,n},c_n)\to(\mu_x,\mu_y,c)$ for covariance.
    For any sequence $k_n\to\infty$ with $k_n/n\to0$, let $\Delta_{f,*}=\lim_n\Delta_f(D_n)$.
    Then, we have the following result:
    \begin{align}
        \liminf_{n\to\infty}\risk^f_{n,k_n}(D_n)\ge\Delta_{f,*}^2\ve.\label{eq:local-minimax-risk-lower-bound}
    \end{align}
\end{theorem}

\begin{proof}[Proof outline]
    Fix $D_n$ and add a multiset $H$ to form auxiliary datasets $D_H=D_n\cup H$ inside the neighborhood.
    We reduce estimation of $f(D_H)$ to estimation of the added records' total contribution $\sum_{z\in H}\psi_{f,D_n}(z)$.
    Eq.~\eqref{eq:moment-influence-sensitivity-coefficients} gives
    \begin{align*}
        n(f(D_H)-f(D_n))
        =\sum_{z\in H}\psi_{f,D_n}(z)+O\!\left(\frac{|H|^2}{n}\right).
    \end{align*}
    Thus, if $|H|^2/n\to0$, any private estimator $A_n$ of $f$ induces the estimator $n(A_n(D_H)-f(D_n))$ of $\sum_{z\in H}\psi_{f,D_n}(z)$.
    From the lower bound on the sum estimation, the MSE of the above estimator is at least $\Delta_f(D_n)^2\ve$.
    Since the MSE of $A_n$ is approximately the same as that of the sum estimator divided by $n^2$,
    the MSE of $A_n(D_H)$ is also lower bounded by $\Delta_f(D_n)^2\ve/n^2$.
    Since $D_H \in B_{k_n}(D_n)$, the worst-case MSE over the neighborhood is also lower bounded by $\Delta_f(D_n)^2\ve/n^2$.
    See~\cref{app:local-lower-bound-proofs} for the full proofs.
\end{proof}

\section{Mean Estimation via Self-Calibration}\label{sec:mean-estimation}
A natural question is whether we can construct a mechanism that achieves this lower bound uniformly over all datasets.
In this and subsequent sections, we answer this question affirmatively by constructing a two-phase mean estimator achieving the error in~\cref{thm:local-minimax-lower-bounds} asymptotically.
Here, we focus on the mean estimation problem to illustrate the main ideas and techniques for achieving instance-optimality.

\subsection{Two-Phase Estimation Procedure}
First, we describe the two-phase estimation procedure with a fixed budget allocation
to illustrate the main idea of the proposed mechanism.
Note that we later introduce an adaptive budget allocation strategy and refine the procedure accordingly.

\paragraph{Phase 1: Preliminary estimation.}
We estimate the mean and count by transformed noise addition with budget $\beta$, using independent $U,V\sim\Lap(1/\beta)$:
\begin{align}
    q(D)      & =b=((n+s(D)/R)/2,(n-s(D)/R)/2),\label{eq:mean-preliminary-query}\displaybreak[1]                                        \\
    \bar s(D) & =s(D)+R(U-V),\quad \bar n=\max\{n+U+V,1\},\quad \bar\mu=\Pi_\mu(\bar s(D)/\bar n).\label{eq:mean-preliminary-estimates}
\end{align}
As shown in the next section, the error of the preliminary estimate affects only higher-order terms in the MSE of the final mean estimator.

\paragraph{Phase 2: Main estimation.}
Given a preliminary estimate $\bar\mu$, we center the dataset around $\bar\mu$, so that
the sum $s(D)-n\bar\mu=\sum_{x\in D}(x-\bar\mu)$ has sensitivity $R+|\bar\mu|$.
Then, we add independent staircase noise to privately release the centered sum and compute the final mean estimate as follows:
\begin{align}
    \begin{aligned}
        W                & =s(D)-n\bar\mu+(R+|\bar\mu|)T_{\varepsilon-\beta},
        \\
        M_{\mu,\beta}(D) & =\Pi_\mu\left(\bar\mu+\frac{W}{\bar n}\right).
    \end{aligned}\label{eq:fixed-budget-mean-estimator}
\end{align}
Here, we reuse the preliminary estimate of the count $\bar n$ to normalize the centered sum.
By adaptive composition~\citep{dwork2014algorithmic}, this estimator satisfies $\varepsilon$-DP.

\subsection{Error Analysis and Budget Allocation}\label{sec:mean-error-budget-allocation}
The error before the final projection is evaluated as
\begin{align*}
    \bar\mu+\frac{W}{\bar n}-\mu
    =\frac{\bar n-n}{\bar n}(\bar\mu-\mu)
    +\frac{R+|\bar\mu|}{\bar n}T_{\varepsilon-\beta}.
\end{align*}
Since the error of the preliminary estimate $\bar\mu - \mu$ appears only in the first term as a product with the relative count error,
the effect on the MSE of the final mean estimator is relatively small.
Specifically, for $0<\beta\le\varepsilon/2$, $n\ge2$, and $n\beta\ge1$, the two-phase estimator $M_{\mu,\beta}$ satisfies
\begin{align}
    n^2L_\mu(M_{\mu,\beta},D)
     & \le(R+|\mu|)^2v_{\varepsilon-\beta}
    +O\left(\frac1{n\beta}+\frac1{n^2\beta^4}+n^2e^{-n\beta/4}\right).\label{eq:fixed-budget-mean-risk-bound}
\end{align}
See~\cref{app:fixed-budget-mean-error} for the derivation.
If we set $\beta$ as
\begin{align*}
    \beta=\min\left\{\frac\varepsilon2,Cn^{-2/5}\right\},
\end{align*}
for some constant $C>0$, we have
\begin{align*}
    \lim_{n\to\infty} n^2L_\mu(M_{\mu,\beta},D)
     & \le(R+|\mu|)^2v_{\varepsilon},
\end{align*}
which matches the local minimax error.
However, since $n$ is not known in advance, we cannot set $\beta$ directly based on $n$.
Thus, we need to select the preliminary budget adaptively.

\subsection{Selecting the Preliminary Budget Privately}\label{sec:private-budget-selection}
To privately select the preliminary budget, we develop an adaptive budget allocation mechanism
based on the gradual release technique~\citep{koufogiannis2015gradual}.
First, we define the budget sequence by $\beta_j=(\varepsilon/2)2^{-j}$ for $j\ge0$.
Then, we generate independent noise sequences for $i=1,\ldots,d$ by
\begin{align}
    Z_{i,0}\sim\Lap(1),\qquad
    Z_{i,j}=\tfrac12Z_{i,j-1}+I_{i,j}\sigma_{i,j}E_{i,j},\quad j\ge1.\label{eq:correlated-laplace-recursion}
\end{align}
Here $I_{i,j}\sim\operatorname{Bernoulli}(3/4)$, $\Prob{\sigma_{i,j}=1}=\Prob{\sigma_{i,j}=-1}=1/2$,
and $E_{i,j}\sim\operatorname{Exp}(1)$.
The initial noises and all variables $I_{i,j},\sigma_{i,j},E_{i,j}$ are mutually independent.
As shown in~\citet{koufogiannis2015gradual}, the marginal distribution of $Z_{i,j}$ is $\Lap(1)$.
For generality, we consider $d$-dimensional queries $q(D)\in\R^d$ with sensitivity at most one and satisfying $q(D)_i \geq 0$ for all $i=1,\ldots,d$ and $\sum_{i=1}^d q(D)_i = |D|$.
Then, we define the noisy query and count estimate by
\begin{align*}
    Y_j=q(D)+\frac{Z_j}{\beta_j},\qquad
    \bar N_j=\sum_{i=1}^dY_{j,i}=|D|+\frac1{\beta_j}\sum_{i=1}^dZ_{i,j}.
\end{align*}
where $Z_j=(Z_{1,j},\ldots,Z_{d,j})$.

Instead of the ideal budget $\beta = C n^{-2/5}$,
we consider the following condition with noisy count estimates:
\begin{align*}
    \bar N_j\ge(C/\beta_j)^{5/2}.
\end{align*}
Then, we select the largest index $j$ whose noisy count meets the above threshold.
That is, the selected budget index is defined as
\begin{align}
    J=\max\bigl(\{0\}\cup\{j\ge1:\bar N_j\ge(C/\beta_j)^{5/2}\}\bigr).\label{eq:budget-selection-index}
\end{align}
As shown in the following lemma, $J$ is finite almost surely
and releasing $(J, \beta_J, Y_J)$ behaves like satisfying $\beta_J$-DP.

\begin{lemma}[Finiteness and privacy of preliminary selection]\label[lemma]{lem:budget-selection-finiteness-privacy}
    The index $J$ in Eq.~\eqref{eq:budget-selection-index} is almost surely finite.
    In addition, for every pair of neighboring datasets $D$ and $E$, every $j\ge0$,
    and every measurable $S\subseteq\R^d$, we have
    \begin{align*}
        \Prob{J=j,\ Y_j\in S\mid D}\le e^{\beta_j}\Prob{J=j,\ Y_j\in S\mid E}.
    \end{align*}
\end{lemma}
\begin{proof}[Proof outline]
    For all sufficiently large $j$, the threshold $(C/\beta_j)^{5/2}$ is at least $2|D|$.
    From a union bound over the $d$ Laplace noise, we have
    \begin{align*}
        \Prob{\bar N_j\ge(C/\beta_j)^{5/2}}
        \le d\exp\left(-\frac{C^{5/2}\beta_j^{-3/2}}{2d}\right).
    \end{align*}
    Since $\beta_j^{-3/2}$ grows geometrically, we obtain
    \begin{align*}
        \sum_{j=1}^\infty\Prob{\bar N_j\ge(C/\beta_j)^{5/2}}<\infty.
    \end{align*}
    By the Borel--Cantelli lemma, only finitely many levels meet the threshold almost surely.
    Therefore, $J$ is almost surely finite.

    From the construction of $Z_{i,j}$, we have
    \begin{align*}
        Y_{k+1,i}=Y_{k,i} + \frac{I_{i,k+1}\sigma_{i,k+1}E_{i,k+1}}{\beta_{k+1}}.
    \end{align*}
    Thus, $Y_{l,i}$ for $l>k$ is obtained from $Y_{k,i}$ by adding independent noise.
    Since the probability of $J = j$ given $Y_j=y$ only depends on the future noise and thresholds,
    $\Prob{J=j\mid Y_j=y}$ does not depend on the dataset.
    Then, we obtain
    \begin{align*}
        \Prob{J=j,\ Y_j\in S\mid D}
        =\int_S\Prob{J=j\mid Y_j=y}\left(\frac{\beta_j}{2}\right)^d
        \exp\bigl(-\beta_j\norm{y-q(D)}_1\bigr)dy.
    \end{align*}
    For neighboring $D,E$, the factor $\Prob{J=j\mid Y_j=y}$ is the same,
    while the ratio of the Laplace densities is bounded by $e^{\beta_j}$.
    This implies the desired guarantee.
    See~\cref{app:mean-privacy-proof} for the full proof.
\end{proof}

\subsubsection{Practical Implementation}\label{sec:budget-selection-finite-implementation}
Computing the maximum in Eq.~\eqref{eq:budget-selection-index} directly is infeasible because it involves infinitely many candidates.
In this section, we develop a practical procedure to compute the maximum in Eq.~\eqref{eq:budget-selection-index} in finite time.

First, for each coordinate $i=1,\ldots,d$, we independently construct a random set $\mathcal S_i$ of levels $j$ where $E_{i,j}$ exceeds the threshold $t_j=-\log(1-e^{-2^{-j}})$.
Specifically, we draw $N_{\mathrm{cand},i}\sim\operatorname{Poisson}(1)$, then draw independent indices $K_{i,r}\sim\operatorname{Geom}(1/2)$ on $\{1,2,\ldots\}$ for $r=1,\ldots,N_{\mathrm{cand},i}$.
The set $\mathcal S_i$ is defined as $\mathcal S_i=\{K_{i,r}:1\le r\le N_{\mathrm{cand},i}\}$.
For each $i$, the counts of occurrences of each level $j$ among the sampled indices $K_{i,r}$ are independent and follow $\operatorname{Poisson}(2^{-j})$ distributions.
That is, we have
\begin{align*}
    \Prob{j\in\mathcal S_i}=1-e^{-2^{-j}}=\Prob{E_{i,j}>t_j}.
\end{align*}

Next, conditional on $\mathcal S_1,\ldots,\mathcal S_d$, we generate the exponential variables $E_{i,j}$ independently across $(i,j)$
using the following procedure: if $j\in\mathcal S_i$, draw $E'_{i,j}\sim\operatorname{Exp}(1)$ and set $E_{i,j}=t_j+E'_{i,j}$;
otherwise, draw $E_{i,j}$ from the exponential distribution conditioned on $E_{i,j}\le t_j$.
The variables $E_{i,j}$ constructed above are mutually independent and follow $\operatorname{Exp}(1)$ distributions.
See~\cref{prop:conditional-exponential-sampling} for the proof.

Let $j_*=\max(\{0\}\cup\bigcup_{i=1}^d\mathcal S_i)$
be the last index where any coordinate exceeds its threshold.
From the construction of $Z$, we have that for all $j>j_*$,
\begin{align*}
    |Z_{i,j}|
     & \le 2^{-(j-j_*)}|Z_{i,j_*}|+\sum_{k=j_*+1}^j2^{-(j-k)}t_k \\
     & \le 2^{-(j-j_*)}|Z_{i,j_*}|+2(j+1)\log2.
\end{align*}
Thus, the estimated count $\bar N_j$ satisfies
\begin{align*}
    \beta_j\bar N_j
     & \le \beta_j|D|+2^{-(j-j_*)}\sum_{i=1}^d|Z_{i,j_*}|+2d(j+1)\log2
    =:U_j.
\end{align*}
Define the stopping level by
\begin{align*}
    j_{\mathrm{stop}}=\min\bigl\{j>j_*:U_j<C^{5/2}\beta_j^{-3/2}\bigr\}.
\end{align*}
For every $j\ge j_{\mathrm{stop}}$, it holds that
\begin{align*}
    \frac{U_j}{\beta_j} \leq 2^{2(j-j_{\mathrm{stop}})}\frac{U_{j_{\mathrm{stop}}}}{\beta_{j_{\mathrm{stop}}}} <2^{2(j-j_{\mathrm{stop}})}\left(\frac{C}{\beta_{j_{\mathrm{stop}}}}\right)^{5/2} = 2^{2(j-j_{\mathrm{stop}})} \left(\frac{C}{2^{(j-j_{\mathrm{stop}})}\beta_{j}}\right)^{5/2} \leq \left(\frac{C}{\beta_j}\right)^{5/2}.
\end{align*}
since $U_{j+1}\le2U_j$ and $\beta_{j+1}=\beta_j/2$.
Here, the second inequality follows from the definition of $j_{\mathrm{stop}}$.
From the definition of $U_j$, we have $\bar N_j \leq \frac{U_j}{\beta_j}$.
Combining the above inequalities, we obtain $\bar N_j < (C/\beta_j)^{5/2}$ for all $j \ge j_{\mathrm{stop}}$,
which implies $J < j_{\mathrm{stop}}$.
Thus, we do not need to generate any further candidate levels beyond $j_{\mathrm{stop}}$.
See Algorithm~\ref{alg:private-budget-selection} for the detailed procedure.

\begin{algorithm}[htbp]
    \caption{Adaptive Budget Allocation}\label{alg:private-budget-selection}
    \small
    \begin{algorithmic}[1]
        \REQUIRE Dataset $D\in\cD_f$, query $q:\cD_f\to\R^d$, parameters $\varepsilon,C>0$.
        \STATE Set $\beta_j=(\varepsilon/2)2^{-j}$ for $j\ge0$, and $t_j=-\log(1-e^{-2^{-j}})$ for $j\ge1$.
        \FOR{$i=1,\ldots,d$}
        \STATE Draw $Z_{i,0}\sim\Lap(1)$ and $N_{\mathrm{cand},i}\sim\operatorname{Poisson}(1)$ independently.
        \STATE Draw $K_{i,r}\sim\operatorname{Geom}(1/2)$ independently for $r=1,\ldots,N_{\mathrm{cand},i}$.
        \STATE Set $\mathcal S_i=\{K_{i,r}:1\le r\le N_{\mathrm{cand},i}\}$.
        \ENDFOR
        \STATE Set $j_*=\max(\{0\}\cup\bigcup_{i=1}^d\mathcal S_i)$.
        \STATE Set $Z_0=(Z_{1,0},\ldots,Z_{d,0})$ and $(J,Y)=(0,q(D)+Z_0/\beta_0)$.
        \FOR{$j=1,\ldots,j_*$}
        \FOR{$i=1,\ldots,d$}
        \IF{$j\in\mathcal S_i$}
        \STATE Draw $E'_{i,j}\sim\operatorname{Exp}(1)$ and set $E_{i,j}=t_j+E'_{i,j}$.
        \ELSE
        \STATE Draw $E_{i,j}$ from $\operatorname{Exp}(1)$ conditioned on $E_{i,j}\le t_j$.
        \ENDIF
        \STATE Draw $I_{i,j}\sim\operatorname{Bernoulli}(3/4)$ and $\sigma_{i,j}\sim\operatorname{Unif}(\{-1,1\})$ independently.
        \STATE Set $Z_{i,j}=\tfrac12Z_{i,j-1}+I_{i,j}\sigma_{i,j}E_{i,j}$.
        \ENDFOR
        \STATE Set $Z_j=(Z_{1,j},\ldots,Z_{d,j})$, $Y_j=q(D)+Z_j/\beta_j$, and $\bar N_j=|D|+\beta_j^{-1}\sum_{i=1}^dZ_{i,j}$.
        \IF{$\bar N_j\ge(C/\beta_j)^{5/2}$}
        \STATE Set $(J,Y)=(j,Y_j)$.
        \ENDIF
        \ENDFOR
        \STATE Define $U_j=\beta_j|D|+2^{-(j-j_*)}\sum_{i=1}^d|Z_{i,j_*}|+2d(j+1)\log2$ for $j\ge j_*$.
        \STATE Set $j_{\mathrm{stop}}=j_*+1$.
        \WHILE{$U_{j_{\mathrm{stop}}}\ge C^{5/2}\beta_{j_{\mathrm{stop}}}^{-3/2}$}
        \STATE Set $j_{\mathrm{stop}}=j_{\mathrm{stop}}+1$.
        \ENDWHILE
        \FOR{$j=j_*+1,\ldots,j_{\mathrm{stop}}-1$}
        \STATE Generate $Z_j,Y_j$ by the same conditional sampling and recursion as above.
        \IF{$\sum_{i=1}^dY_{j,i}\ge(C/\beta_j)^{5/2}$}
        \STATE Set $(J,Y)=(j,Y_j)$.
        \ENDIF
        \ENDFOR
        \RETURN $(J,Y,\beta_J)$.
    \end{algorithmic}
\end{algorithm}

\subsection{Mean Estimation with Private Budget Selection}\label{sec:private-budget-mean-estimation}
For mean estimation, we apply~\cref{alg:private-budget-selection} to the two-coordinate query in Eq.~\eqref{eq:mean-preliminary-query}.
From its output $(J,Y_J,\beta)$, where $\beta=\beta_J$, we reconstruct the preliminary mean $\bar\mu$ and count $\bar n$ as in Eq.~\eqref{eq:mean-preliminary-estimates}.
Using the remaining budget $\varepsilon-\beta$, we release a noisy sum of $x-\bar\mu$ and reconstruct the mean as
\begin{align}
    \begin{aligned}
        W        & =\sum_{x\in D}(x-\bar\mu)+\Delta_0T_{\varepsilon-\beta},
        \\
        M_\mu(D) & =\Pi_\mu\left(\bar\mu+\frac{W}{\bar n}\right).
    \end{aligned}\label{eq:private-budget-mean-estimator}
\end{align}
Here $\Delta_0=R+|\bar\mu|$ is the sensitivity of the sum for the fixed preliminary output.
\Cref{alg:mean-estimation} gives the full procedure.

\begin{algorithm}[htbp]
    \caption{Mean Estimation ($M_\mu$)}\label{alg:mean-estimation}
    \small
    \begin{algorithmic}[1]
        \REQUIRE Dataset $D\in\cD_\mu$; parameters $R,\varepsilon,C>0$.
        \STATE \textbf{Phase 1: Preliminary estimation.}
        \STATE Set $q(D)=\sum_{x\in D}\bigl((R+x)/(2R),(R-x)/(2R)\bigr)$.
        \STATE Call~\cref{alg:private-budget-selection} with $(D,q,\varepsilon,C)$ to obtain $(J,Y_J,\beta)$.
        \STATE Set $\bar n=\max\{\sum_{i=1}^2Y_{J,i},1\}$.
        \STATE Set $\bar\mu=\Pi_\mu\bigl(R(Y_{J,1}-Y_{J,2})/\bar n\bigr)$.
        \STATE \textbf{Phase 2: Main estimation.}
        \STATE Set $r_0(x)=x-\bar\mu$.
        \STATE Set $\Delta_0=R+|\bar\mu|$.
        \STATE Conditional on $(J,Y_J)$, draw fresh $T\sim T_{\varepsilon-\beta}$.
        \STATE Set $W=\sum_{x\in D}r_0(x)+\Delta_0T$.
        \RETURN $\Pi_\mu(\bar\mu+W/\bar n)$.
    \end{algorithmic}
\end{algorithm}

As shown in the following proposition, the mechanism satisfies $\varepsilon$-DP.
\begin{proposition}[Privacy]\label[proposition]{prop:mean-estimator-privacy}
    $M_\mu$ satisfies $\varepsilon$-DP.
\end{proposition}
\begin{proof}[Proof outline]
    From~\cref{lem:budget-selection-finiteness-privacy} and the properties of the staircase mechanism,
    we have
    \begin{align*}
        \Prob{J=j,Y_J\in dy\mid D}
         & \le e^{\beta_j}\Prob{J=j,Y_J\in dy\mid E},                 \\
        \Prob{W\in dw\mid J=j,Y_J=y,D}
         & \le e^{\varepsilon-\beta_j}\Prob{W\in dw\mid J=j,Y_J=y,E}.
    \end{align*}
    Multiplying these bounds gives
    \begin{align*}
         & \Prob{J=j,Y_J\in dy,W\in dw\mid D}                                            \\
         & \quad\le e^{\beta_j}e^{\varepsilon-\beta_j}\Prob{J=j,Y_J\in dy,W\in dw\mid E} \\
         & \quad=e^\varepsilon\Prob{J=j,Y_J\in dy,W\in dw\mid E}.
    \end{align*}
    Thus, releasing $(J,Y_J,W)$ satisfies $\varepsilon$-DP.
    From the post-processing property, we obtain the result.
    See~\cref{app:mean-privacy-proof} for the full proof.
\end{proof}

In addition, the mechanism achieves the local minimax rate for mean estimation.
\begin{theorem}[Optimality for mean estimation]\label{thm:mean-local-minimax-optimality}
    If $|D_n|=n$, $\mu(D_n)\to\mu\in(-R,R)$, and $k_n\to\infty$ with $k_n/n\to0$,
    $M_\mu$ satisfies
    \begin{align}
        \limsup_{n\to\infty}\sup_{E\in\Ball{k_n}{D_n}}n^2L_\mu(M_\mu,E) & \le (R+|\mu|)^2\ve,\label{eq:mean-local-risk-attainment}
    \end{align}
    This matches the lower bound in~\cref{thm:local-minimax-lower-bounds}.
\end{theorem}
See~\cref{app:mean-uniform-risk-bound} for the proof.

\paragraph{Comparison with Transformed Noise Addition.}
Define the optimal asymptotic coefficient as $C_*(\mu)=(R+|\mu|)^2\ve$.
Comparing the above bound with Eq.~\eqref{eq:hourglass-coefficient} yields
\begin{align*}
    \frac{C_*(\mu)}{C_{\mathrm{H}}(\mu)} & =\frac{(R+|\mu|)^2}{2(R^2+\mu^2)} \leq 1.
\end{align*}
The ratio is $1/2$ at zero and approaches one at an endpoint.
Since data distribute widely across the domain, the empirical mean is likely to be far from the endpoints $-R$ and $R$.
Therefore, the improvement around 0 is particularly relevant in practice.

\section{Variance and Covariance Estimation}\label{sec:variance-covariance-estimation}
In this section, we extend the two-phase mean estimator to variance and covariance.

First, we call~\cref{alg:private-budget-selection} with the B\'ezier query $q_f(D)=b_f$ in~\cref{sec:bezier} for $f\in\{v,c\}$,
using $d_v=3$ for variance and $d_c=4$ for covariance.
Then, we obtain the output $(J,Y_J,\beta)$, where $\beta=\beta_J$.

For variance, we compute preliminary estimates of the mean $\bar\mu$, variance $\bar v$, and count $\bar n$ from $Y_J$.
Using the remaining budget $\varepsilon-\beta$, we release a noisy sum of $(x-\bar\mu)^2-\bar v$ and reconstruct the variance as
\begin{align}
    \begin{aligned}
        W      & =\sum_{x\in D}\bigl((x-\bar\mu)^2-\bar v\bigr)+\Delta_0T_{\varepsilon-\beta}, \\
        M_v(D) & =\Pi_v\left(\bar v+\frac{W}{\bar n}\right).
    \end{aligned}\label{eq:variance-estimator}
\end{align}
Here $\Delta_0=\max\{\bar v,(R+|\bar\mu|)^2-\bar v\}$ is the sensitivity of the sum for the fixed preliminary output.
\Cref{alg:variance-estimation} gives the full procedure.

\begin{algorithm}[htbp]
    \caption{Variance Estimation ($M_v$)}\label{alg:variance-estimation}
    \small
    \begin{algorithmic}[1]
        \REQUIRE Dataset $D\in\cD_v$; parameters $R,\varepsilon,C>0$.
        \STATE \textbf{Phase 1: Preliminary estimation.}
        \STATE Set $q(D)=\sum_{x\in D}b_v(x)$, with $b_v(x)$ defined in~\cref{sec:bezier}.
        \STATE Call~\cref{alg:private-budget-selection} with $(D,q,\varepsilon,C)$ to obtain $(J,Y_J,\beta)$.
        \STATE Set $\bar n=\max\{\sum_{i=1}^3Y_{J,i},1\}$.
        \STATE Set $\bar\mu=\Pi_\mu\bigl(2R(Y_{J,2}/2+Y_{J,3})/\bar n-R\bigr)$.
        \STATE Set $\bar v=\Pi_v\bigl(4R^2Y_{J,3}/\bar n-(\bar\mu+R)^2\bigr)$.
        \STATE \textbf{Phase 2: Main estimation.}
        \STATE Set $r_0(x)=(x-\bar\mu)^2-\bar v$.
        \STATE Set $\Delta_0=\max\{\bar v,(R+|\bar\mu|)^2-\bar v\}$.
        \STATE Conditional on $(J,Y_J)$, draw fresh $T\sim T_{\varepsilon-\beta}$.
        \STATE Set $W=\sum_{x\in D}r_0(x)+\Delta_0T$.
        \RETURN $\Pi_v(\bar v+W/\bar n)$.
    \end{algorithmic}
\end{algorithm}

For covariance, we compute preliminary estimates of the two means $\bar\mu_x,\bar\mu_y$, covariance $\bar c$, and count $\bar n$ from $Y_J$.
Using the remaining budget $\varepsilon-\beta$, we release a noisy sum of $(x-\bar\mu_x)(y-\bar\mu_y)-\bar c$ and reconstruct the covariance as
\begin{align}
    \begin{aligned}
        W      & =\sum_{(x,y)\in D}\bigl((x-\bar\mu_x)(y-\bar\mu_y)-\bar c\bigr)+\Delta_0T_{\varepsilon-\beta}, \\
        M_c(D) & =\Pi_c\left(\bar c+\frac{W}{\bar n}\right).
    \end{aligned}\label{eq:covariance-estimator}
\end{align}
Here $\Delta_0=\max_{x,y\in\{-R,R\}}|(x-\bar\mu_x)(y-\bar\mu_y)-\bar c|$ is the sensitivity of the sum for the fixed preliminary output.
\Cref{alg:covariance-estimation} gives the full procedure.

\begin{algorithm}[htbp]
    \caption{Covariance Estimation ($M_c$)}\label{alg:covariance-estimation}
    \small
    \begin{algorithmic}[1]
        \REQUIRE Dataset $D\in\cD_c$; parameters $R,\varepsilon,C>0$.
        \STATE \textbf{Phase 1: Preliminary estimation.}
        \STATE Set $q(D)=\sum_{(x,y)\in D}b_c(x,y)$, with $b_c(x,y)$ defined in~\cref{sec:bezier}.
        \STATE Call~\cref{alg:private-budget-selection} with $(D,q,\varepsilon,C)$ to obtain $(J,Y_J,\beta)$.
        \STATE Set $\bar n=\max\{\sum_{i=1}^4Y_{J,i},1\}$.
        \STATE Set $\bar\mu_x=\Pi_\mu\bigl(2R(Y_{J,3}+Y_{J,4})/\bar n-R\bigr)$.
        \STATE Set $\bar\mu_y=\Pi_\mu\bigl(2R(Y_{J,2}+Y_{J,4})/\bar n-R\bigr)$.
        \STATE Set $\bar c=\Pi_c\bigl(4R^2Y_{J,4}/\bar n-(\bar\mu_x+R)(\bar\mu_y+R)\bigr)$.
        \STATE \textbf{Phase 2: Main estimation.}
        \STATE Set $r_0(x,y)=(x-\bar\mu_x)(y-\bar\mu_y)-\bar c$.
        \STATE Set $\Delta_0=\max_{x,y\in\{-R,R\}}|r_0(x,y)|$.
        \STATE Conditional on $(J,Y_J)$, draw fresh $T\sim T_{\varepsilon-\beta}$.
        \STATE Set $W=\sum_{(x,y)\in D}r_0(x,y)+\Delta_0T$.
        \RETURN $\Pi_c(\bar c+W/\bar n)$.
    \end{algorithmic}
\end{algorithm}

As in mean estimation, both mechanisms satisfy $\varepsilon$-DP and achieve the local minimax rate asymptotically.
\begin{proposition}[Privacy]\label[proposition]{prop:variance-covariance-privacy}
    $M_f$ satisfies $\varepsilon$-DP for $f\in\{v,c\}$.
\end{proposition}

\begin{theorem}[Optimality for variance and covariance estimation]\label{thm:variance-covariance-local-minimax-optimality}
    Let $f\in\{v,c\}$.
    If $|D_n|=n$, $(\mu_n,v_n)\to(\mu,v)$ for variance or $(\mu_{x,n},\mu_{y,n},c_n)\to(\mu_x,\mu_y,c)$ for covariance, and $k_n\to\infty$ with $k_n/n\to0$,
    $M_f$ satisfies
    \begin{align*}
        \limsup_{n\to\infty}\sup_{E\in\Ball{k_n}{D_n}}n^2L_f(M_f,E)\le\Delta_{f,*}^2\ve,
    \end{align*}
    where $\Delta_{f,*}=\lim_n\Delta_f(D_n)$. This matches the lower bound in~\cref{thm:local-minimax-lower-bounds}.
\end{theorem}
See~\cref{app:variance-covariance-privacy-utility,app:variance-covariance-error-analysis} for the proofs.

The same mechanisms also attain the asymptotic global minimax error.
\begin{corollary}[Global minimax optimality]\label[corollary]{cor:variance-covariance-global-minimax-optimality}
    For $f\in\{v,c\}$, the estimator $M_f$ satisfies
    \begin{align*}
        \lim_{n\to\infty}\sup_{\substack{D\in\cD_f \\|D|\ge n}}|D|^2L_f(M_f,D)
        =16R^4\ve,
    \end{align*}
    which matches the lower bound in~\citet{takakura2025optimal}.
\end{corollary}
See~\cref{app:variance-covariance-global-minimax-proof} for the proof.
The global minimax error and a mechanism attaining it were previously known only in the high-privacy regime.
The above corollary extends this characterization to all privacy regimes, including the low-privacy regime.
This result would be of independent interest.

\paragraph{Comparison with B\'ezier Mechanism.}
Define the optimal asymptotic coefficients as $C_{*,f}(D)=\Delta_f(D)^2\ve$ for $f\in\{v,c\}$.
Comparing them with the B\'ezier coefficients in Eq.~\eqref{eq:bezier-coefficients} gives
\begin{align}
    \begin{aligned}
        \frac{\varepsilon^2\ve}{8}
         & \le\frac{C_{*,c}(D)}{C_{\mathrm{B},c}(D)}\le\frac{\varepsilon^2\ve}{2} \leq 1, \\
        \frac{\varepsilon^2\ve}{22}
         & \le\frac{C_{*,v}(D)}{C_{\mathrm{B},v}(D)}\le\frac{\varepsilon^2\ve}{2} \leq 1.
    \end{aligned}\label{eq:variance-covariance-bezier-risk-ratios}
\end{align}
In the high-privacy limit $\varepsilon\downarrow0$, we have $\varepsilon^2\ve/2\to1$, and the bounds reduce to
\begin{align*}
    \frac14\le\lim_{\varepsilon\downarrow0}\frac{C_{*,c}(D)}{C_{\mathrm{B},c}(D)}\le1,\qquad
    \frac1{11}\le\lim_{\varepsilon\downarrow0}\frac{C_{*,v}(D)}{C_{\mathrm{B},v}(D)}\le1.
\end{align*}
The lower bounds are attained at $\mu_x=\mu_y=0,c=0$, and $\mu=0, v=R^2/2$, respectively.

\section{Numerical Experiments}\label{sec:experiments}
In this section, we compare the proposed estimators with the hourglass mechanism for the mean and
the B\'ezier mechanism for variance and covariance, using synthetic datasets with $R=1$.
Each condition uses $300{,}000$ trials per method to estimate the MSE.
Error bars show the standard error.
See~\cref{app:experiments} for the detailed experimental setup.

\subsection{Mean estimation.}
We examine how the MSE approaches the sharp asymptotic coefficient as the count grows and how the improvement over the hourglass mechanism varies with the mean location.

\begin{figure}[htbp]
    \centering
    \includegraphics[width=\textwidth]{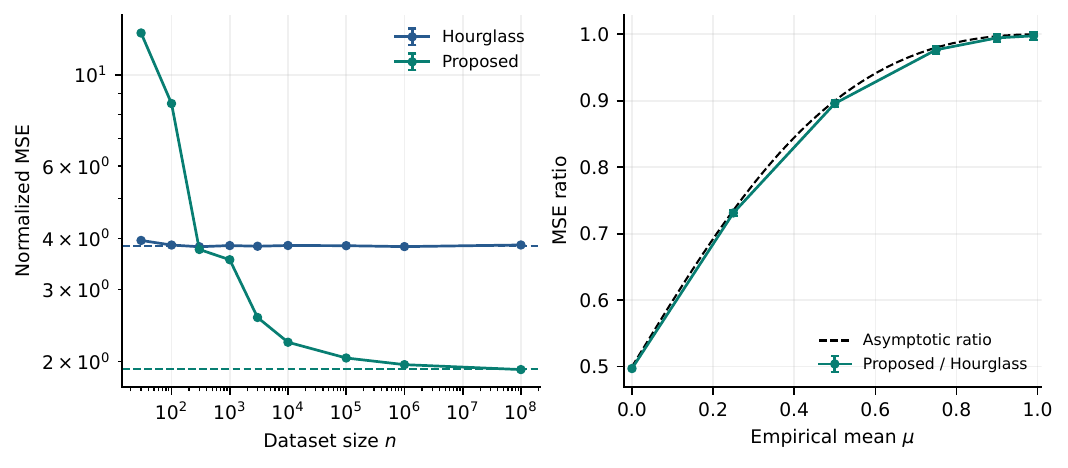}
    \caption{Left: normalized MSE with varying dataset size. Right: MSE ratios to hourglass mechanism with varying $\mu$. Dashed lines show asymptotic coefficients or ratios.}\label{fig:mean-estimation-performance}
\end{figure}

The left panel of~\Cref{fig:mean-estimation-performance} shows that the normalized MSE approaches the sharp asymptotic coefficient as the count grows.
In this experiment, we fix $\mu = 0, \varepsilon = 1$ and vary the count $n$ from $30$ to $10^8$.
For small datasets, preliminary estimation can increase the error, but the effect diminishes as the count grows.
At large counts, the proposed estimator achieves approximately half the MSE of the hourglass mechanism.

The right panel of~\Cref{fig:mean-estimation-performance} plots the ratio of the proposed estimator's MSE
to that of the hourglass mechanism against the empirical mean $\mu$.
In this experiment, we set $n = 10^8, \varepsilon = 1$ and vary the empirical mean $\mu$ across the domain.
We observe that the improvement is more pronounced when the mean is near the center of the domain and diminishes
as the mean moves towards the boundary.

\subsection{Variance and covariance estimation.}
\begin{figure}[htbp]
    \centering
    \includegraphics[width=\textwidth]{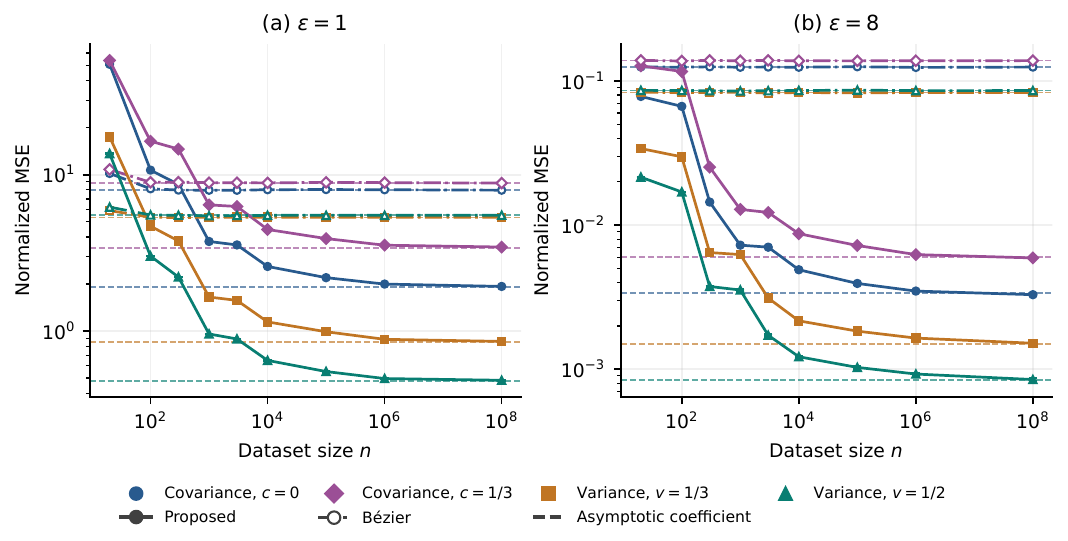}
    \caption{Normalized MSE for variance and covariance estimation with $\varepsilon=1$ (left) and $\varepsilon=8$ (right). Dashed lines show asymptotic coefficients.}\label{fig:variance-covariance-performance}
\end{figure}

For variance and covariance estimation,
we evaluate the performance of the proposed mechanisms across different privacy budgets and dataset sizes.
We consider four centered datasets: two with covariances $c=0$ and $c=1/3$ and two with variances $v=1/3$ and $v=1/2$.
\Cref{fig:variance-covariance-performance} illustrates the performance of the proposed estimators and existing B\'ezier estimators for these datasets
for $\varepsilon=1$ (left panel) and $\varepsilon=8$ (right panel).
We observe that the normalized MSE of the proposed estimators approaches the optimal MSE as the dataset size increases, outperforming the B\'ezier estimators.
The improvement is larger at $\varepsilon=8$, consistent with the privacy-dependent factor in Eq.~\eqref{eq:variance-covariance-bezier-risk-ratios}.
Additional results for other dataset configurations are reported in~\cref{app:nonzero-moment-results}.

\FloatBarrier
\section{Conclusion}
In this paper, we studied mean, variance, and covariance estimation under pure differential privacy in the add-remove model.
To measure the performance of private estimators beyond worst-case analysis, we consider the local minimax error and provide sharp asymptotic lower bounds for it.
Then, we proposed two-stage estimation procedures by developing self-calibration and adaptive budget allocation techniques.
We prove that they asymptotically achieve the local minimax error and outperform existing mechanisms uniformly across different datasets.
Our experimental results validate the theoretical findings, showing that the performance of the proposed estimators approaches the optimal error as the dataset size increases.
\label{sec:main-end}

\clearpage
\begingroup
\small
\setlength{\bibsep}{1pt}
\bibliographystyle{plainnat}
\bibliography{references}
\endgroup
\clearpage
\appendix

\section{Asymptotic Error of Transformed Noise Addition}\label{app:transformed-noise-error-analysis}
Here, we provide the detailed derivations of the asymptotic MSE for transformed noise addition.

\subsection{Case 1: Laplace Noise}\label{app:transformed-laplace-error-analysis}
\begin{proof}[Proof of the Laplace MSE bound]
    Let $Z_1,Z_2\sim\Lap(1/\varepsilon)$ and
    \begin{align*}
        B=Z_1+Z_2,\qquad \xi_n=(R-\mu_n)Z_1-(R+\mu_n)Z_2.
    \end{align*}
    Then, the error of the unprojected estimator is given by
    \begin{align*}
        \frac{n\mu_n+R(Z_1-Z_2)}{n+B}-\mu_n=\frac{\xi_n}{n+B}.
    \end{align*}
    Since the variance of $Z_i$ is $2/\varepsilon^2$, we obtain
    \begin{align*}
        \Expec{\xi_n^2}
        =\frac{2\big((R-\mu_n)^2+(R+\mu_n)^2\big)}{\varepsilon^2}
        =\frac{4(R^2+\mu_n^2)}{\varepsilon^2}.
    \end{align*}

    Let $G_n=\{|B|\le n/2\}$.
    A tail bound on Laplace noise gives $\Prob{G_n^c}\le Ke^{-an}$ for some constants $K,a>0$ depending only on $\varepsilon$.
    On $G_n$, $n/(n+B)\le2$ and $|n^2/(n+B)^2-1|\le10|B|/n$.
    Thus, we arrive at
    \begin{align*}
        \Expec{\frac{n^2\xi_n^2}{(n+B)^2}\1_{G_n}}=\Expec{\xi_n^2}+O(n^{-1}).
    \end{align*}
    Since projection does not increase MSE, we have the following upper bound:
    \begin{align*}
        L_\mu(\hat\mu_{\mathrm{T}},D_n)\le\frac{4(R^2+\mu_n^2)}{\varepsilon^2n^2}+o(n^{-2}).
    \end{align*}
\end{proof}

\subsection{Case 2: Hourglass Noise}\label{app:transformed-hourglass-error-analysis}
To improve the utility of transformed noise addition with Laplace noise,~\citet{kulesza2024mean} introduced hourglass noise.
\begin{definition}[Hourglass noise~\citep{kulesza2024mean}]\label[definition]{def:hourglass-noise}
    Fix $\varepsilon>0$, and define $\gamma_\varepsilon$ as in~\cref{sec:laplace-and-staircase-mechanisms}.
    The distribution $(Z_1,Z_2)\sim\operatorname{Hourglass}(\varepsilon)$
    is defined as $Z_1$ following staircase distribution with density $f_\varepsilon$ and the conditional probabilities for $Z_2$ given $Z_1$ as follows:
    \begin{align*}
        \Prob{Z_2=k-z\mid Z_1=z}
        =\frac{1-q_\varepsilon}{1+q_\varepsilon}q_\varepsilon^{\left|k-\operatorname{sgn}(z)\floor{|z|+1-\gamma_\varepsilon}\right|},\qquad k\in\mathbb Z.
    \end{align*}
\end{definition}

\begin{proof}[Proof of the Hourglass MSE bound]
    Let $(Z_1, Z_2)\sim\operatorname{Hourglass}(\varepsilon)$.
    As shown in~\citet{kulesza2024mean}, $Z_i$ have mean zero, variance $\ve$.
    Thus the numerator $\xi_n$ defined in the proof for Laplace noise satisfies
    \begin{align*}
        \Expec{\xi_n^2}
        =\big((R-\mu_n)^2+(R+\mu_n)^2\big)\ve
        =2(R^2+\mu_n^2)\ve.
    \end{align*}
    As in the proof for Laplace noise, we obtain the result.
\end{proof}

\section{Asymptotic Error of B\'ezier Mechanisms}\label{app:bezier-error-analysis}
Here, we derive the asymptotic MSE in Eq.~\eqref{eq:bezier-coefficients} for variance and covariance estimation.

\subsection{Variance Estimation}\label{app:bezier-variance-error-analysis}
\begin{proof}[Proof of the variance MSE bound]
    Let $Z=(Z_1,Z_2,Z_3)$ consist of the independent $\Lap(1/\varepsilon)$ noises added to the B\'ezier vector.
    For $t=(t_1,t_2,t_3)$ with $\sum_{j=1}^3t_j>-1$, define
    \begin{align*}
        f_v(t)=\frac{\mu^2+v+R^2(t_1-t_2+t_3)}{1+\sum_{j=1}^3t_j}
        -\left(\frac{\mu+R(t_3-t_1)}{1+\sum_{j=1}^3t_j}\right)^2.
    \end{align*}

    Lemma A.4 of~\citet{takakura2025optimal} implies that
    \begin{align*}
        n^2 L_v(\widehat v_{\mathrm{B}},D)\le\Expec{A_v(Z)^2}+o(1),
        \qquad A_v(Z)=\langle\grad f_v(0),Z\rangle.
    \end{align*}
    From the definition of $f_v$, we have
    \begin{align*}
        A_v(Z)
         & =\big((R+\mu)^2-v\big)Z_{1}
        +(\mu^2-R^2-v)Z_{2}+\big((R-\mu)^2-v\big)Z_{3}.
    \end{align*}
    Since the $Z_{j}$ are independent with mean zero and variance $2/\varepsilon^2$, we obtain
    \begin{align*}
        \Expec{A_v(Z)^2}
         & =\frac2{\varepsilon^2}\left[\big((R+\mu)^2-v\big)^2
        +(\mu^2-R^2-v)^2+\big((R-\mu)^2-v\big)^2\right]                     \\
         & =\frac2{\varepsilon^2}\big[3(\mu^2-v)^2+2R^2(5\mu^2-v)+3R^4\big]
        =C_{\mathrm{B},v}.
    \end{align*}
\end{proof}

\subsection{Covariance Estimation}\label{app:bezier-covariance-error-analysis}
\begin{proof}[Proof of the covariance MSE bound]
    Let $Z=(Z_1,\ldots,Z_4)$ consist of the independent $\Lap(1/\varepsilon)$ noises added to the B\'ezier vector.
    For $t=(t_1,\ldots,t_4)$ with $\sum_{j=1}^4t_j>-1$, define
    \begin{align*}
        f_c(t)
         & =\frac{\mu_x\mu_y+c+R^2(t_1-t_2-t_3+t_4)}{1+\sum_{j=1}^4t_j} \\
         & \quad-\frac{\mu_x+R(-t_1-t_2+t_3+t_4)}{1+\sum_{j=1}^4t_j}
        \frac{\mu_y+R(-t_1+t_2-t_3+t_4)}{1+\sum_{j=1}^4t_j}.
    \end{align*}
    Lemma A.4 of~\citet{takakura2025optimal} implies that
    \begin{align*}
        n^2 L_c(\widehat c_{\mathrm{B}},D)\le\Expec{A_c(Z)^2}+o(1),
        \qquad A_c(Z)=\langle\grad f_c(0),Z\rangle.
    \end{align*}
    From the definition of $f_c$, we have
    \begin{align*}
        A_c(Z)
         & =\big((R+\mu_x)(R+\mu_y)-c\big)Z_1
        -\big((R+\mu_x)(R-\mu_y)+c\big)Z_2         \\
         & \quad-\big((R-\mu_x)(R+\mu_y)+c\big)Z_3
        +\big((R-\mu_x)(R-\mu_y)-c\big)Z_4.
    \end{align*}
    Since the $Z_{j}$ are independent with mean zero and variance $2/\varepsilon^2$, we obtain
    \begin{align*}
        \Expec{A_c(Z)^2}
         & =\frac2{\varepsilon^2}\Big[\big((R+\mu_x)(R+\mu_y)-c\big)^2
        +\big((R+\mu_x)(R-\mu_y)+c\big)^2                                            \\
         & \qquad+\big((R-\mu_x)(R+\mu_y)+c\big)^2
        +\big((R-\mu_x)(R-\mu_y)-c\big)^2\Big]                                       \\
         & =\frac8{\varepsilon^2}\big[R^4+R^2(\mu_x^2+\mu_y^2)+(\mu_x\mu_y-c)^2\big]
        =C_{\mathrm{B},c}.
    \end{align*}
\end{proof}

\section{Proofs of the Local Lower Bounds}\label{app:local-lower-bound-proofs}
We prove~\cref{thm:local-minimax-lower-bounds} by reduction to sum estimation.
We first establish the sum-estimation lower bound in Lemma 5.1 of~\citet{kulesza2024mean}.
\begin{lemma}[Lower bound for sum estimation]\label[lemma]{lem:sum-estimation-minimax-lower-bound}
    Fix $\varepsilon>0$.
    Let $\cU_m$ be the collection of multisets in $[0,1]$ containing at most $m$ records, and write $s(U)=\sum_{t\in U}t$.
    Then,
    \begin{align}
        \inf_{\widehat s\in\cA_\varepsilon^s}\sup_{U\in\cU_m}\Expec{(\widehat s(U)-s(U))^2}
        \ge\ve(1-o(1)),\qquad m\to\infty.\label{eq:sum-estimation-minimax-lower-bound}
    \end{align}
\end{lemma}

\begin{proof}
    We prove by contradiction.
    If the bound does not hold, there exist a constant $\rho<\ve$ and,
    along a sequence $m_i\to\infty$, $\varepsilon$-DP estimators $M_i$ such that
    \begin{align*}
        \sup_{U\in\cU_{m_i}}\Expec{(M_i(U)-s(U))^2}\le\rho.
    \end{align*}
    Fix $i \geq 1$ and define $d=\ell=\floor{\sqrt{m_i}}$ and $t_j=j/d$ for $j=1,\ldots,d$.
    For each $h\in\cH=\{0,\ldots,\ell-1\}^{d}$,
    we consider the dataset $U_h$ containing $h_j$ copies of $t_j$ for each $j$.
    That is, $U_h$ may contain $d$ values $t_1,\ldots,t_d$ and each value $t_j$ appears at most $\ell-1$ times.
    Then $|U_h|\le d(\ell-1)\le m_i$.

    Let $P_h$ be the law of $M_i(U_h)-s(U_h)$, and define
    \begin{align*}
        Q=\frac{1}{|\cH|}\sum_{h\in\cH}P_h.
    \end{align*}
    From the assumption, we have $\int z^2Q(dz)\le\rho$.

    If $h_j\le\ell-2$, the datasets $U_h$ and $U_{h+e_j}$ differ by one record, where $e_j$ denotes the $j$-th standard basis vector.
    Since $s(U_{h+e_j})=s(U_h)+t_j$, $\varepsilon$-DP implies
    \begin{align*}
        P_h(A)\le e^\varepsilon P_{h+e_j}(A-t_j),
        \qquad P_{h+e_j}(A)\le e^\varepsilon P_h(A+t_j)
    \end{align*}
    for any measurable set $A$.
    Averaging the first inequality over $\cH$, we obtain
    \begin{align*}
        Q(A)
         & =\frac1{|\cH|}\sum_{\substack{h\in\cH                 \\h_j\le\ell-2}}P_h(A)
        +\frac1{|\cH|}\sum_{\substack{h\in\cH                    \\h_j=\ell-1}}P_h(A)\\
         & \le\frac{e^\varepsilon}{|\cH|}\sum_{\substack{h\in\cH \\h_j\le\ell-2}}P_{h+e_j}(A-t_j)+\frac{\ell^{d-1}}{\ell^d}\\
         & =\frac{e^\varepsilon}{|\cH|}\sum_{\substack{g\in\cH   \\g_j\ge1}}P_g(A-t_j)+\frac1\ell
        \le e^\varepsilon Q(A-t_j)+\frac1\ell.
    \end{align*}
    Similarly, averaging the second inequality over $\cH$, we obtain
    \begin{align*}
        Q(A)
         & =\frac1{|\cH|}\sum_{\substack{g\in\cH                 \\g_j\ge1}}P_g(A)
        +\frac1{|\cH|}\sum_{\substack{g\in\cH                    \\g_j=0}}P_g(A)\\
         & \le\frac{e^\varepsilon}{|\cH|}\sum_{\substack{g\in\cH \\g_j\ge1}}P_{g-e_j}(A+t_j)+\frac1\ell
        \le e^\varepsilon Q(A+t_j)+\frac1\ell.
    \end{align*}
    Thus, we arrive at
    \begin{align}
        Q(A)\le e^\varepsilon Q(A\pm t_j)+\frac1{\ell}.
        \label{eq:sum-averaged-histogram-privacy}
    \end{align}
    As $i\to\infty$, $\ell$ also goes to infinity and we obtain
    \begin{align*}
        Q_\infty(A)\le e^\varepsilon Q_\infty(A+t),
    \end{align*}
    where $Q_\infty$ is the limit of $Q$ as $i\to\infty$.
    Then, Theorems 4 and 7 in \citet{geng2015optimal} yield
    \begin{align*}
        \ve\le\int z^2Q_\infty(dz)\le\liminf_{i\to\infty}\int z^2Q(dz)\le\rho,
    \end{align*}
    contradicting $\rho<\ve$. This completes the proof.
\end{proof}

\subsection{Lower Bound for Mean Estimation}\label{app:mean-local-lower-bound}
\begin{proof}
    Without loss of generality, we may assume $\mu_n=\mu(D_n)\ge0$ for every $n$ and write $\Delta_{\mu,n}=\Delta_\mu(D_n)=R+\mu_n$.
    For $t\in[0,1]$, set $x_n(t)=\mu_n-\Delta_{\mu,n}t\in[-R,R]$.
    Given any finite multiset $U$ in $[0,1]$, define
    \begin{align*}
        E_n(U)=D_n\cup\{x_n(t):t\in U\}.
    \end{align*}
    For any $A_n\in\cA_\varepsilon$, construct the sum estimator
    \begin{align*}
        \widehat s_n(U)=\frac{n}{\Delta_{\mu,n}}(\mu_n-A_n(E_n(U))).
    \end{align*}
    Note that $\widehat s_n$ satisfies $\varepsilon$-DP since $D_n$ is fixed and $E_n(U)$ changes by at most one record when $U$ changes by one record.
    Let $m_n=\min\{k_n,\floor{n^{1/3}}\}$, and thus $m_n\to\infty$, $m_n\le k_n$, and $m_n^2/n\to0$.
    Fix $U\in\cU_{m_n}$ and write $r=|U|$.
    By the definition of the empirical mean, we obtain
    \begin{align*}
        \widehat s_n(U)-s(U)
         & =\frac{n}{\Delta_{\mu,n}}\bigl(\mu(E_n(U))-A_n(E_n(U))\bigr)                    \\
         & \quad+\left\{\frac{n}{\Delta_{\mu,n}}\bigl(\mu_n-\mu(E_n(U))\bigr)-s(U)\right\} \\
         & =\frac{n}{\Delta_{\mu,n}}\bigl(\mu(E_n(U))-A_n(E_n(U))\bigr)-\frac{rs(U)}{n+r}.
    \end{align*}
    The first term has $L^2$ norm $n\sqrt{L_\mu(A_n,E_n(U))}/\Delta_{\mu,n}$, while $0\le rs(U)/(n+r)\le m_n^2/n=o(1)$ since $0\le s(U)\le r\le m_n$.
    Applying~\cref{lem:sum-estimation-minimax-lower-bound} and the triangle inequality in $L^2$ to the decomposition above gives
    \begin{align*}
        \sqrt{\ve}-o(1)
         & \le\sup_{U\in\cU_{m_n}}\sqrt{\Expec{(\widehat s_n(U)-s(U))^2}}                            \\
         & \le\frac{n}{\Delta_{\mu,n}}\sup_{E\in\Ball{k_n}{D_n}}\sqrt{L_\mu(A_n,E)}+\frac{m_n^2}{n}.
    \end{align*}
    Thus, taking the infimum over all $A_n\in\cA_\varepsilon$ gives
    \begin{align*}
        \sqrt{\risk^\mu_{n,k_n}(D_n)}\ge\Delta_{\mu,n}(\sqrt{\ve}-o(1)).
    \end{align*}
    Since $\Delta_{\mu,n}\to R+|\mu|$, we obtain the result.
\end{proof}

\subsection{Lower Bound for Variance and Covariance Estimation}\label{app:variance-covariance-local-lower-bounds}
\begin{proof}[Variance and covariance cases of~\cref{thm:local-minimax-lower-bounds}]
    Write $\Delta_{f,n}=\Delta_f(D_n)$.
    From the definitions of empirical variance and covariance, we have
    \begin{align*}
        \frac1n\sum_{x\in D_n}\psi_{v,D_n}(x)
         & =\frac1n\sum_{x\in D_n}(x-\mu_n)^2-v_n=0,                    \\
        \frac1n\sum_{(x,y)\in D_n}\psi_{c,D_n}(x,y)
         & =\frac1n\sum_{(x,y)\in D_n}(x-\mu_{x,n})(y-\mu_{y,n})-c_n=0.
    \end{align*}
    Thus, $\psi_{f,D_n}$ takes both a nonpositive and a nonnegative value on $D_n$.
    Continuity implies that its image is a closed interval containing zero.
    Since $\Delta_{f,n}=\max_z|\psi_{f,D_n}(z)|$, one endpoint is $\Delta_{f,n}$ or $-\Delta_{f,n}$.
    Hence, the image of $\psi_{f,D_n}$ contains $[0,\Delta_{f,n}]$ or $[-\Delta_{f,n},0]$.
    Thus, we can choose a sign $s_n\in\{-1,1\}$ and records $z_n(t)$ such that
    \begin{align*}
        s_n\psi_{f,D_n}(z_n(t))=\Delta_{f,n}t
    \end{align*}
    for any $t\in[0,1]$.

    For any finite multiset $U$ in $[0,1]$, set $E_n(U)=D_n\cup\{z_n(t):t\in U\}$.
    Given any $A_n\in\cA_\varepsilon$, define
    \begin{align*}
        \widehat s_n(U)=\frac{s_n n}{\Delta_{f,n}}(A_n(E_n(U))-f(D_n))
    \end{align*}
    As in the proof of mean estimation, it satisfies $\varepsilon$-DP.
    Let $m_n=\min\{k_n,\floor{n^{1/3}}\}$, and thus $m_n\to\infty$, $m_n\le k_n$, and $m_n^2/n\to0$.

    Fix $U\in\cU_{m_n}$ and write $r=|U|$.
    We decompose the sum estimation error as
    \begin{align*}
        \widehat s_n(U)-s(U)
         & =\frac{s_n n}{\Delta_{f,n}}\bigl(A_n(E_n(U))-f(E_n(U))\bigr)                      \\
         & \quad+\left\{\frac{s_n n}{\Delta_{f,n}}\bigl(f(E_n(U))-f(D_n)\bigr)-s(U)\right\}.
    \end{align*}
    The first term has $L^2$ norm $n\sqrt{L_f(A_n,E_n(U))}/\Delta_{f,n}$.
    To bound the second term, for covariance, write $z_n(t)=(x_n(t),y_n(t))$ and set
    \begin{align*}
        a_x=\sum_{t\in U}(x_n(t)-\mu_{x,n}),\qquad
        a_y=\sum_{t\in U}(y_n(t)-\mu_{y,n}).
    \end{align*}
    For variance, set $y_n(t)=x_n(t)$ and $\mu_{x,n}=\mu_{y,n}=\mu_n$.
    Using $s_n\psi_{f,D_n}(z_n(t))=\Delta_{f,n}t$ and expanding the empirical covariance, we obtain
    \begin{align*}
        \left|\frac{s_n n}{\Delta_{f,n}}(f(E_n(U))-f(D_n))-s(U)\right|
         & =\frac1{\Delta_{f,n}}\left|n(f(E_n(U))-f(D_n))-\sum_{t\in U}\psi_{f,D_n}(z_n(t))\right|                   \\
         & =\frac1{\Delta_{f,n}}\left|-\frac{r}{n+r}\sum_{t\in U}\psi_{f,D_n}(z_n(t))-\frac{na_xa_y}{(n+r)^2}\right| \\
         & \le\frac1{\Delta_{f,n}}\left(\frac{5R^2r^2}{n+r}+\frac{4nR^2r^2}{(n+r)^2}\right)                          \\
         & \le\frac{9R^2r^2}{n\Delta_{f,n}}
        \le\frac{18m_n^2}{n}=o(1).
    \end{align*}
    Here, $|\psi_{f,D_n}(z)|\le5R^2$, $|a_x|,|a_y|\le2Rr$, $r\le m_n$, and $\Delta_{f,n}\ge R^2/2$.
    Applying~\cref{lem:sum-estimation-minimax-lower-bound} and the triangle inequality in $L^2$ to the decomposition above, we obtain
    \begin{align*}
        \sqrt{\ve}-o(1)
         & \le\sup_{U\in\cU_{m_n}}\sqrt{\Expec{(\widehat s_n(U)-s(U))^2}}                          \\
         & \le\frac{n}{\Delta_{f,n}}\sup_{E\in\Ball{k_n}{D_n}}\sqrt{L_f(A_n,E)}+\frac{18m_n^2}{n}.
    \end{align*}
    Taking the infimum over $A_n$ and using $\Delta_{f,n}\to\Delta_{f,*}$ completes the proof.
\end{proof}

\section{Proofs for Mean Estimation}\label{app:mean-proofs}
In this section, we provide proofs for the mean estimation results, including adaptive budget allocation.
\subsection{Fixed-Budget Error Analysis}\label{app:fixed-budget-mean-error}
\begin{proof}[Proof of Eq.~\eqref{eq:fixed-budget-mean-risk-bound}]
    Let $n\ge2$ and $0<\beta\le\varepsilon/2$ satisfy $n\beta\ge1$.
    Write $B=U+V$ and $\mathcal G=\{|B|\le n/2\}$.
    As in~\cref{sec:mean-error-budget-allocation}, substituting $s(D)=n\mu$ into Eq.~\eqref{eq:fixed-budget-mean-estimator} yields
    \begin{align*}
        \bar\mu+\frac W{\bar n}-\mu
         & =\frac{\bar n-n}{\bar n}(\bar\mu-\mu)+\frac{R+|\bar\mu|}{\bar n}T_{\varepsilon-\beta}
        \\
         & =\frac{B(\bar\mu-\mu)}{n+B}+\frac{R+|\bar\mu|}{n+B}T_{\varepsilon-\beta}.
    \end{align*}
    Hence,
    \begin{align*}
        n^2L_\mu(M_{\mu,\beta},D)
         & \le n^2\Expec{\left(\bar\mu+\frac W{\bar n}-\mu\right)^2\1_{\mathcal G}}+4R^2n^2\Prob{\mathcal G^c}
        \\
         & =v_{\varepsilon-\beta}\Expec{\frac{n^2(R+|\bar\mu|)^2}{(n+B)^2}\1_{\mathcal G}}
        \\
         & \quad+\Expec{\frac{n^2B^2(\bar\mu-\mu)^2}{(n+B)^2}\1_{\mathcal G}}+4R^2n^2\Prob{\mathcal G^c}.
    \end{align*}
    To bound the first term, note that projection and $n+B\ge n/2$ on $\mathcal G$ give, for $p\in\{1,4\}$,
    \begin{align*}
        \Expec{|\bar\mu-\mu|^p\1_{\mathcal G}}
         & \le\Expec{\left|\frac{R(U-V)-\mu B}{n+B}\right|^p\1_{\mathcal G}}
        \\
         & \le\left(\frac{2R}n\right)^p\Expec{(|U-V|+|B|)^p}
        =O((n\beta)^{-p}).
    \end{align*}
    Since $|n^2/(n+B)^2-1|\le16|B|/n$ on $\mathcal G$ and $|(R+|\bar\mu|)^2-(R+|\mu|)^2|\le4R|\bar\mu-\mu|$, we have
    \begin{align*}
        \Expec{\frac{n^2(R+|\bar\mu|)^2}{(n+B)^2}\1_{\mathcal G}}
         & \le(R+|\mu|)^2+\frac{64R^2}{n}\Expec{|B|}
        \\
         & \quad+4R\Expec{|\bar\mu-\mu|\1_{\mathcal G}}
        \\
         & =(R+|\mu|)^2+O((n\beta)^{-1}).
    \end{align*}
    For the second term, $n^2/(n+B)^2\le4$ on $\mathcal G$ and the Cauchy--Schwarz inequality yield
    \begin{align*}
        \Expec{\frac{n^2B^2(\bar\mu-\mu)^2}{(n+B)^2}\1_{\mathcal G}}
         & \le4\bigl(\Expec{B^4}\Expec{(\bar\mu-\mu)^4\1_{\mathcal G}}\bigr)^{1/2}
        \\
         & =O(n^{-2}\beta^{-4}).
    \end{align*}
    Finally, $\Prob{\mathcal G^c}\le\Prob{|U|>n/4}+\Prob{|V|>n/4}\le2e^{-n\beta/4}$.
    Substituting these bounds into the MSE decomposition and using the boundedness of $v_{\varepsilon-\beta}$ for $\varepsilon-\beta\in[\varepsilon/2,\varepsilon]$, we conclude that
    \begin{align*}
        n^2L_\mu(M_{\mu,\beta},D)
        \le(R+|\mu|)^2v_{\varepsilon-\beta}
        +O\left(\frac1{n\beta}+\frac1{n^2\beta^4}+n^2e^{-n\beta/4}\right).
    \end{align*}
\end{proof}

\subsection{Adaptive-Budget Privacy Analysis}\label{app:mean-privacy-proof}
\begin{proof}[Proof of~\cref{lem:budget-selection-finiteness-privacy}]
    For all sufficiently large $j$, the threshold $(C/\beta_j)^{5/2}$ is at least $2|D|$.
    From a union bound, we have
    \begin{align*}
        \Prob{\bar N_j\ge(C/\beta_j)^{5/2}}
         & \le\Prob{\sum_{i=1}^d Z_{i,j}\ge\tfrac12 C^{5/2}\beta_j^{-3/2}}
        \\
         & \le d\exp\left(-\frac{C^{5/2}\beta_j^{-3/2}}{2d}\right).
    \end{align*}
    Note that the marginal distributions of $Z_{i,j}$ are Laplace distribution with the scale parameter 1.
    Since $\beta_j^{-3/2}$ grows geometrically, these probabilities are summable.
    The Borel--Cantelli lemma implies that only finitely many levels meet the threshold almost surely.
    Thus, $J$ is almost surely finite.

    For every $j\ge0$ and measurable set $S\subseteq\R^d$, conditioning on $Y_j$ yields
    \begin{align*}
        \Prob{J=j,\ Y_J\in S\mid D}
        =\int_S\Prob{J=j\mid Y_j=y,D}\left(\frac{\beta_j}{2}\right)^d\exp\bigl(-\beta_j\norm{y-q(D)}_1\bigr)dy.
    \end{align*}
    From the definition of $J$ in Eq.~\eqref{eq:budget-selection-index}, we have
    \begin{align*}
        \{J=j\}
         & =\{\bar N_j\ge(C/\beta_j)^{5/2}\}\cap\bigcap_{k>j}\{\bar N_k<(C/\beta_k)^{5/2}\},\qquad j\ge1,
        \\
        \{J=0\}
         & =\bigcap_{k\ge1}\{\bar N_k<(C/\beta_k)^{5/2}\}.
    \end{align*}
    Given $Y_j=y$, the current count $\bar N_j=\sum_i y_i$ is fixed, and the recursion above generates all later outputs from increments independent of $D$ and $Y_j$.
    Therefore, the conditional selection probability satisfies
    \begin{align*}
        \Prob{J=j\mid Y_j=y,D}=\Prob{J=j\mid Y_j=y},\qquad j\ge0,
    \end{align*}
    where the right-hand side does not depend on $D$.

    For neighboring $D,E$, the ratio of the Laplace densities is bounded by $e^{\beta_j}$.
    Substituting this bound into the integral above yields
    \begin{align}
        \begin{aligned}
             & \Prob{J=j,\ Y_J\in S\mid D}
            \\
             & \quad\le e^{\beta_j}\int_S\Prob{J=j\mid Y_j=y}\left(\frac{\beta_j}{2}\right)^d\exp\bigl(-\beta_j\norm{y-q(E)}_1\bigr)dy
            \\
             & \quad=e^{\beta_j}\Prob{J=j,\ Y_J\in S\mid E}.
        \end{aligned}\label{eq:selected-preliminary-output-privacy}
    \end{align}
    On $\{J=j\}$, we have $Y_J=Y_j$, which completes the proof.
\end{proof}
\begin{proof}[Proof of~\cref{prop:mean-estimator-privacy}]
    Let $D,E$ be neighboring datasets.
    Fix the preliminary output $(J,Y_J)=(j,y)$.
    Note that adding or removing one record changes $\sum_{x\in D}(x-\bar\mu)$ by at most $R+|\bar\mu|$.
    From~\cref{lem:budget-selection-finiteness-privacy} and the staircase mechanism's privacy guarantee, we have
    \begin{align*}
        \Prob{J=j,Y_J\in dy\mid D}
         & \le e^{\beta_j}\Prob{J=j,Y_J\in dy\mid E},
        \\
        \Prob{W\in dw\mid J=j,Y_J=y,D}
         & \le e^{\varepsilon-\beta_j}\Prob{W\in dw\mid J=j,Y_J=y,E}.
    \end{align*}
    Multiplying these bounds, we obtain
    \begin{align*}
         & \Prob{J=j,Y_J\in dy,W\in dw\mid D}
        \\
         & \quad\le e^{\beta_j}e^{\varepsilon-\beta_j}\Prob{J=j,Y_J\in dy,W\in dw\mid E}
        \\
         & \quad=e^\varepsilon\Prob{J=j,Y_J\in dy,W\in dw\mid E}.
    \end{align*}
    Integrating over $(j, y,w)$ yields
    \begin{align}
        \Prob{(J,Y_J,W)\in B\mid D}
        \le e^\varepsilon\Prob{(J,Y_J,W)\in B\mid E}\label{eq:selected-budget-privacy-composition}
    \end{align}
    for any measurable set $B$.
    Therefore, the joint release $(J,Y_J,W)$ satisfies $\varepsilon$-DP.
    Then, the post-processing property proves the result.
\end{proof}

\subsection{Equivalence of Conditional Sampling}\label{app:conditional-exponential-sampling}
\begin{proposition}\label[proposition]{prop:conditional-exponential-sampling}
    Let $(E_{i,j})_{1\le i\le d,\ j\ge1}$ be generated by the conditional sampling procedure in~\cref{sec:budget-selection-finite-implementation}.
    Then, these variables are mutually independent and each follows the $\operatorname{Exp}(1)$ distribution.
\end{proposition}
\begin{proof}[Proof of~\cref{prop:conditional-exponential-sampling}]
    Since the construction is independent across coordinates,
    it suffices to show the claim for coordinate $i$.
    Define
    \begin{align*}
        N_j=\sum_{r=1}^{N_{\mathrm{cand},i}}\1_{\{K_{i,r}=j\}},\qquad j\ge1.
    \end{align*}
    Since $N_{\mathrm{cand},i}\sim\operatorname{Poisson}(1)$ and $\Prob{K_{i,r}=j}=2^{-j}$, Poisson splitting~\citep{last2018lectures} implies that the $N_j$ are independent with $N_j\sim\operatorname{Poisson}(2^{-j})$.
    Hence, the events $\{j\in\mathcal S_i\}=\{N_j>0\}$ are independent and have probabilities $1-e^{-2^{-j}}=e^{-t_j}$.

    Given $\mathcal S_i$, the $E_{i,j}$ are independent and have the $\operatorname{Exp}(1)$ law conditioned on $E_{i,j}>t_j$ for $j\in\mathcal S_i$ and on $E_{i,j}\le t_j$ otherwise.
    Thus, the marginal distribution of $E_{i,j}$ can be computed as
    \begin{align*}
        \Prob{E_{i,j}\in A}
         & =(1-e^{-t_j})\frac{\int_{A\cap[0,t_j]}e^{-u}du}{1-e^{-t_j}}+e^{-t_j}\frac{\int_{A\cap(t_j,\infty)}e^{-u}du}{e^{-t_j}} \\
         & =\int_{A} e^{-u}du
    \end{align*}
    for any measurable set $A \subset [0, \infty)$.
    Hence, $E_{i,j}\sim\operatorname{Exp}(1)$.
    This completes the proof.
\end{proof}

\subsection{Adaptive-Budget Error Analysis}\label{app:mean-uniform-risk-bound}
\begin{proof}[Proof of~\cref{thm:mean-local-minimax-optimality}]
    For an input $D$ with sufficiently large $n=|D|$, write $\mu=\mu(D)$ and $B_j=(Z_{1,j}+Z_{2,j})/\beta_j$.
    Define
    \begin{align*}
        j_-          & =\max\{j\ge0:(C/\beta_j)^{5/2}\le n/2\}, \\
        j_+          & =\min\{j\ge0:(C/\beta_j)^{5/2}\ge2n\},   \\
        \mathcal I_n & =\{j_-,\ldots,j_+-1\},\qquad
        \mathcal G_n=\{J\in\mathcal I_n,\ |B_J|\le n/2\}.
    \end{align*}
    Since $(C/\beta_j)^{5/2}$ increases by a factor of $2^{5/2}$ at each level, $\mathcal I_n$ has bounded size and $\beta_j=\Theta(n^{-2/5})$ on this set.
    On $\mathcal G_n$, the selected preliminary count satisfies $\bar n=n+B_J\ge n/2$.
    Substituting $s(D)=n\mu$ into the estimator, we obtain
    \begin{align*}
        \bar\mu+\frac W{\bar n}-\mu
         & =\frac{\bar n-n}{\bar n}(\bar\mu-\mu)+\frac{R+|\bar\mu|}{\bar n}T_{\varepsilon-\beta_J}
        \\
         & =\frac{B_J(\bar\mu-\mu)}{n+B_J}+\frac{R+|\bar\mu|}{n+B_J}T_{\varepsilon-\beta_J}.
    \end{align*}
    Conditionally on $(J,Y_J)$, the staircase noise has mean zero and second moment $v_{\varepsilon-\beta_J}$.
    Squaring the error and averaging over $(J,Y_J)$, we have
    \begin{align*}
        n^2L_\mu(M_\mu,D)
         & \le n^2\Expec{\left(\bar\mu+\frac W{\bar n}-\mu\right)^2\1_{\mathcal G_n}}+4R^2n^2\Prob{\mathcal G_n^c}
        \\
         & =\Expec{\frac{n^2(R+|\bar\mu|)^2v_{\varepsilon-\beta_J}}{(n+B_J)^2}\1_{\mathcal G_n}}
        \\
         & \quad+\Expec{\frac{n^2B_J^2(\bar\mu-\mu)^2}{(n+B_J)^2}\1_{\mathcal G_n}}+4R^2n^2\Prob{\mathcal G_n^c}.
    \end{align*}
    To bound the first term, we first evaluate the preliminary errors.
    For every fixed positive integer $r$, summing the fixed-level Laplace moments over $\mathcal I_n$ yields
    \begin{align*}
        \Expec{|B_J|^r\1_{\mathcal G_n}}
         & \le\sum_{j\in\mathcal I_n}\Expec{|B_j|^r}
        =O(n^{2r/5}).
    \end{align*}
    Projection and $n+B_J\ge n/2$ on $\mathcal G_n$ yield
    \begin{align*}
        \Expec{|\bar\mu-\mu|^r\1_{\mathcal G_n}}
         & \le\Expec{\left|\frac{R(Z_{1,J}-Z_{2,J})/\beta_J-\mu B_J}{n+B_J}\right|^r\1_{\mathcal G_n}}
        \\
         & \le\left(\frac{2R}n\right)^r\sum_{j\in\mathcal I_n}\beta_j^{-r}\Expec{(|Z_{1,j}-Z_{2,j}|+|Z_{1,j}+Z_{2,j}|)^r}
        \\
         & =O(n^{-3r/5}).
    \end{align*}
    Since
    \begin{align*}
        \left|\frac{n^2}{(n+B_J)^2}-1\right|\le\frac{16|B_J|}n,\qquad
        |(R+|\bar\mu|)^2-(R+|\mu|)^2|\le4R|\bar\mu-\mu|
    \end{align*}
    on $\mathcal G_n$, these moment bounds imply
    \begin{align*}
        \Expec{\frac{n^2(R+|\bar\mu|)^2}{(n+B_J)^2}\1_{\mathcal G_n}}
         & \le(R+|\mu|)^2+\frac{64R^2}n\Expec{|B_J|\1_{\mathcal G_n}}
        \\
         & \quad+4R\Expec{|\bar\mu-\mu|\1_{\mathcal G_n}}
        \\
         & =(R+|\mu|)^2+O(n^{-3/5}).
    \end{align*}
    Since $\beta_J=O(n^{-2/5})$ on $\mathcal G_n$ and $v_\eta$ is continuously differentiable on $[\varepsilon/2,\varepsilon]$, we have $v_{\varepsilon-\beta_J}=\ve+O(n^{-2/5})$ uniformly on this event.
    Consequently,
    \begin{align*}
        \Expec{\frac{n^2(R+|\bar\mu|)^2v_{\varepsilon-\beta_J}}{(n+B_J)^2}\1_{\mathcal G_n}}
        \le(R+|\mu|)^2\ve+O(n^{-2/5}).
    \end{align*}
    For the second term, $n^2/(n+B_J)^2\le4$ on $\mathcal G_n$ and the Cauchy--Schwarz inequality yield
    \begin{align*}
        \Expec{\frac{n^2B_J^2(\bar\mu-\mu)^2}{(n+B_J)^2}\1_{\mathcal G_n}}
         & \le4\bigl(\Expec{B_J^4\1_{\mathcal G_n}}\Expec{(\bar\mu-\mu)^4\1_{\mathcal G_n}}\bigr)^{1/2}
        \\
         & =O(n^{-2/5}).
    \end{align*}
    Finally, by the definition of $J$ and the Laplace tail bound,
    \begin{align*}
        \Prob{J<j_-}    & \le2e^{-\beta_{j_-}n/4},                         \\
        \Prob{J\ge j_+} & \le\sum_{j\ge j_+}2e^{-C^{5/2}\beta_j^{-3/2}/4}.
    \end{align*}
    Both bounds are $O(e^{-a n^{3/5}})$ for some $a>0$ by the geometric growth of $\beta_j^{-3/2}$.
    Taking a union bound, we obtain
    \begin{align}
        \begin{aligned}
            \Prob{\mathcal G_n^c}
             & \le\Prob{J<j_-}+\Prob{J\ge j_+}+\sum_{j\in\mathcal I_n}\Prob{|B_j|>n/2}
            \\
             & \le\Prob{J<j_-}+\Prob{J\ge j_+}+\sum_{j\in\mathcal I_n}2e^{-n\beta_j/4}
            =O(e^{-a n^{3/5}}).
        \end{aligned}\label{eq:mean-selected-good-event-tail}
    \end{align}
    Substituting these bounds into the MSE decomposition, we conclude that
    \begin{align}
        \begin{aligned}
            n^2L_\mu(M_\mu,D)
             & \le(R+|\mu|)^2\ve+O(n^{-2/5})+O(n^2e^{-a n^{3/5}})
            \\
             & =(R+|\mu|)^2\ve+o(1),
        \end{aligned}\label{eq:adaptive-budget-mean-risk-bound}
    \end{align}
    uniformly over $\mu\in[-R,R]$.

    Let $\mu_n=\mu(D_n)$.
    For $E\in\Ball{k_n}{D_n}$, write $n_E=|E|$ and $\mu_E=\mu(E)$.
    We apply Eq.~\eqref{eq:adaptive-budget-mean-risk-bound} to each such $E$.
    Since $E$ is obtained by at most $k_n$ additions or removals, we have
    \begin{align}
        n-k_n\le n_E\le n+k_n,\qquad
        |\mu_E-\mu_n|\le\frac{2Rk_n}{n-k_n}.\label{eq:mean-neighborhood-stability}
    \end{align}
    Therefore, $n_E/n\to1$ and $\mu_E\to\mu$ uniformly over the neighborhood.

    Using Eq.~\eqref{eq:mean-neighborhood-stability} in Eq.~\eqref{eq:adaptive-budget-mean-risk-bound}, we obtain
    \begin{align*}
        \sup_{E\in\Ball{k_n}{D_n}}n^2L_\mu(M_\mu,E)
         & \le\sup_{E\in\Ball{k_n}{D_n}}\left(\frac n{n_E}\right)^2\bigl((R+|\mu_E|)^2\ve+o(1)\bigr)
        \\
         & =(R+|\mu|)^2\ve+o(1).
    \end{align*}
    This completes the proof.
\end{proof}

\section{Proofs for Variance and Covariance Estimation}\label{app:variance-covariance-proofs}
Here, we provide privacy and utility analysis for the proposed estimators of variance and covariance.

\subsection{Privacy Analysis}\label{app:variance-covariance-privacy-utility}
\begin{proof}[Proof of~\cref{prop:variance-covariance-privacy}]
    The B\'ezier basis is nonnegative and sums to one for each record.
    Therefore, the aggregate query $q_f$ has sensitivity one and coordinate sum $|D|$.
    By~\cref{lem:budget-selection-finiteness-privacy}, the selected output satisfies Eq.~\eqref{eq:selected-preliminary-output-privacy} with $d=d_f$.
    Given $(J,Y_J)=(j,y)$, the residual query has sensitivity $\Delta_0$ and its staircase release satisfies $(\varepsilon-\beta_j)$-DP.
    Applying Eq.~\eqref{eq:selected-budget-privacy-composition} in $d_f$ coordinates proves that the joint release satisfies $\varepsilon$-DP.
    The post-processing property proves the result.
\end{proof}

\subsection{Error Analysis}\label{app:variance-covariance-error-analysis}
\begin{proof}[Proof of~\cref{thm:variance-covariance-local-minimax-optimality}]
    Fix $R,\varepsilon,C>0$ and $f\in\{v,c\}$.
    Write $q_f$ for the B\'ezier query used in~\cref{alg:variance-estimation,alg:covariance-estimation}.
    For an input $D$ with sufficiently large $n=|D|$, denote the preliminary estimate of $f(D)$ by $\bar f$.
    Define $h=(\mu_x-\bar\mu_x)(\mu_y-\bar\mu_y)$, $B_j=\beta_j^{-1}\sum_{i=1}^{d_f}Z_{i,j}$ and
    \begin{align*}
        j_-          & =\max\{j\ge0:(C/\beta_j)^{5/2}\le n/2\}, \\
        j_+          & =\min\{j\ge0:(C/\beta_j)^{5/2}\ge2n\},   \\
        \mathcal I_n & =\{j_-,\ldots,j_+-1\},\qquad
        \mathcal G_n=\{J\in\mathcal I_n,\ |B_J|\le n/2\}.
    \end{align*}
    The threshold increases by a factor of $2^{5/2}$ at each level.
    Hence, $\mathcal I_n$ has bounded size and $\beta_j=\Theta(n^{-2/5})$ on this set.
    On $\mathcal G_n$, we have $\bar n=n+B_J\ge n/2$.
    Expanding the estimator, we obtain
    \begin{align*}
        \bar f+\frac W{\bar n}-f(D)
         & =\bar f+\frac{n(f(D)-\bar f+h)+\Delta_0T_{\varepsilon-\beta_J}}{\bar n}-f(D)      \\
         & =\frac{B_J(\bar f-f(D))+nh}{n+B_J}+\frac{\Delta_0}{n+B_J}T_{\varepsilon-\beta_J}.
    \end{align*}
    Conditionally on $(J,Y_J)$, the staircase noise has mean zero and second moment $v_{\varepsilon-\beta_J}$.
    Squaring the error and averaging over $(J,Y_J)$, we have
    \begin{align*}
        n^2L_f(M_f,D)
         & \le\Expec{\frac{n^2\Delta_0^2v_{\varepsilon-\beta_J}}{(n+B_J)^2}\1_{\mathcal G_n}} \\
         & \quad+\Expec{\frac{n^2\{B_J(\bar f-f(D))+nh\}^2}{(n+B_J)^2}\1_{\mathcal G_n}}
        +4R^4n^2\Prob{\mathcal G_n^c}.
    \end{align*}
    To bound the first term, we evaluate the selected preliminary errors.
    For each fixed positive integer $p$, the fixed-level Laplace moments give
    \begin{align*}
        \Expec{\|Z_J/\beta_J\|_1^p\1_{\mathcal G_n}}
        \le\sum_{j\in\mathcal I_n}\Expec{\|Z_j/\beta_j\|_1^p}
        =O(n^{2p/5}).
    \end{align*}
    Each coordinate ratio satisfies, on $\mathcal G_n$,
    \begin{align*}
        \left|\frac{q_{f,i}(D)+Z_{i,J}/\beta_J}{n+B_J}-\frac{q_{f,i}(D)}n\right|
        =\frac{|nZ_{i,J}/\beta_J-q_{f,i}(D)B_J|}{n(n+B_J)}
        \le\frac2n(|Z_{i,J}|/\beta_J+|B_J|).
    \end{align*}
    The reconstruction formulas combine these ratios with products of projected means, which are Lipschitz on $[-R,R]^2$.
    Since projection cannot increase the error, we obtain, for a constant $K_R$ depending only on $R$,
    \begin{align*}
        \bigl(|\bar\mu_x-\mu_x|+|\bar\mu_y-\mu_y|+|\bar f-f(D)|\bigr)\1_{\mathcal G_n}
        \le\frac{K_R}n\|Z_J/\beta_J\|_1\1_{\mathcal G_n}.
    \end{align*}
    Therefore,
    \begin{align*}
        \Expec{|B_J|^p\1_{\mathcal G_n}} & =O(n^{2p/5}),  \\
        \Expec{\bigl(|\bar\mu_x-\mu_x|^p+|\bar\mu_y-\mu_y|^p+|\bar f-f(D)|^p\bigr)\1_{\mathcal G_n}}
                                         & =O(n^{-3p/5}).
    \end{align*}
    Since $\Delta_0$ is bounded and Lipschitz in the projected preliminary moments, we have
    \begin{align*}
        \Expec{|\Delta_0-\Delta_f(D)|\1_{\mathcal G_n}}=O(n^{-3/5}).
    \end{align*}
    Using $|n^2/(n+B_J)^2-1|\le16|B_J|/n$ on $\mathcal G_n$, we obtain
    \begin{align*}
        \Expec{\frac{n^2\Delta_0^2}{(n+B_J)^2}\1_{\mathcal G_n}}
         & =\Expec{\Delta_0^2\1_{\mathcal G_n}}+O\left(\frac{\Expec{|B_J|\1_{\mathcal G_n}}}n\right) \\
         & \le\Delta_f(D)^2+O(n^{-3/5}).
    \end{align*}
    The smoothness of $v_\eta$ on $[\varepsilon/2,\varepsilon]$ and $\beta_J=O(n^{-2/5})$ on $\mathcal G_n$ imply $v_{\varepsilon-\beta_J}=\ve+O(n^{-2/5})$ uniformly on this event.
    Consequently,
    \begin{align*}
        \Expec{\frac{n^2\Delta_0^2v_{\varepsilon-\beta_J}}{(n+B_J)^2}\1_{\mathcal G_n}}
        \le\Delta_f(D)^2\ve+O(n^{-2/5}).
    \end{align*}
    For the second term, the denominator bound and the Cauchy--Schwarz inequality yield
    \begin{align*}
        \Expec{\frac{n^2\{B_J(\bar f-f(D))+nh\}^2}{(n+B_J)^2}\1_{\mathcal G_n}}
         & \le8\bigl(\Expec{B_J^4\1_{\mathcal G_n}}\Expec{(\bar f-f(D))^4\1_{\mathcal G_n}}\bigr)^{1/2}
        +8n^2\Expec{h^2\1_{\mathcal G_n}}                                                               \\
         & =O(n^{-2/5}).
    \end{align*}
    Here, $\Expec{h^2\1_{\mathcal G_n}}\le(\Expec{(\bar\mu_x-\mu_x)^4\1_{\mathcal G_n}}\Expec{(\bar\mu_y-\mu_y)^4\1_{\mathcal G_n}})^{1/2}=O(n^{-12/5})$.
    Finally, a union bound over the $d_f$ coordinates gives $\Prob{|B_j|>t}\le d_fe^{-\beta_jt/d_f}$.
    Applying this bound to the threshold tests, we have
    \begin{align*}
        \Prob{J<j_-}    & \le d_fe^{-\beta_{j_-}n/(2d_f)},                        \\
        \Prob{J\ge j_+} & \le\sum_{j\ge j_+}d_fe^{-C^{5/2}\beta_j^{-3/2}/(2d_f)}.
    \end{align*}
    The geometric growth of $\beta_j^{-3/2}$ makes both bounds $O(e^{-a n^{3/5}})$ for some $a>0$.
    Taking a union bound, we obtain
    \begin{align}
        \begin{aligned}
            \Prob{\mathcal G_n^c}
             & \le\Prob{J<j_-}+\Prob{J\ge j_+}+\sum_{j\in\mathcal I_n}\Prob{|B_j|>n/2}        \\
             & \le\Prob{J<j_-}+\Prob{J\ge j_+}+\sum_{j\in\mathcal I_n}d_fe^{-n\beta_j/(2d_f)}
            =O(e^{-a n^{3/5}}).
        \end{aligned}\label{eq:second-moment-selected-good-event-tail}
    \end{align}
    Substituting these bounds into the MSE decomposition yields
    \begin{align}
        \begin{aligned}
            n^2L_f(M_f,D)
             & \le\Delta_f(D)^2\ve+O(n^{-2/5})+O(n^2e^{-a n^{3/5}}) \\
             & =\Delta_f(D)^2\ve+o(1),
        \end{aligned}\label{eq:adaptive-budget-second-moment-risk-bound}
    \end{align}
    uniformly over inputs of count $n$.

    For $E\in\Ball{k_n}{D_n}$, write $n_E=|E|$.
    Since at most $k_n$ additions or removals change each empirical moment by $O(k_n/n)$, we have $n_E/n\to1$ and $\Delta_f(E)\to\Delta_{f,*}$ uniformly over the neighborhood.
    Applying Eq.~\eqref{eq:adaptive-budget-second-moment-risk-bound} to each $E$, we obtain
    \begin{align*}
        \sup_{E\in\Ball{k_n}{D_n}}n^2L_f(M_f,E)
        \le\sup_{E\in\Ball{k_n}{D_n}}\left(\frac n{n_E}\right)^2\bigl(\Delta_f(E)^2\ve+o(1)\bigr)
        =\Delta_{f,*}^2\ve+o(1).
    \end{align*}
    This completes the proof.
\end{proof}

\subsection{Global Minimax Risk}\label{app:variance-covariance-global-minimax-proof}
\begin{proof}[Proof of~\cref{cor:variance-covariance-global-minimax-optimality}]
    We first bound the sensitivity coefficients.
    For variance, $0\le v\le R^2$ and $|\mu|\le R$ imply $\Delta_v\le4R^2$.
    For covariance, the Cauchy--Schwarz inequality gives $|c|\le R^2-(\mu_x^2+\mu_y^2)/2$.
    Hence,
    \begin{align*}
        \Delta_c
         & \le(R+|\mu_x|)(R+|\mu_y|)+|c|                         \\
         & \le2R^2+R(|\mu_x|+|\mu_y|)-\frac12(|\mu_x|-|\mu_y|)^2
        \le4R^2.
    \end{align*}

    Suppose, for a contradiction, that the required asymptotic upper bound does not hold.
    Then there exist $\delta>0$ and datasets $(D_j)_{j\ge1}$ with $m_j=|D_j|\to\infty$ such that
    \begin{align*}
        m_j^2L_f(M_f,D_j)>16R^4\ve+\delta.
    \end{align*}
    The moment sequence $(\mu(D_j),v(D_j))_{j\ge1}$ for variance, or $(\mu_x(D_j),\mu_y(D_j),c(D_j))_{j\ge1}$ for covariance, is bounded.
    Passing to a convergent subsequence and relabeling the corresponding datasets as $(D_j)_{j\ge1}$, we may assume that these moments converge.
    By~\cref{thm:variance-covariance-local-minimax-optimality} with $k_j=\lfloor\sqrt{m_j}\rfloor$ and $D_j\in\Ball{k_j}{D_j}$, we have
    \begin{align*}
        \limsup_{j\to\infty}m_j^2L_f(M_f,D_j)
        \le\Delta_{f,*}^2\ve\le16R^4\ve,
    \end{align*}
    contradicting the choice of $(D_j)_{j\ge1}$.
    This proves the upper bound.

    Rescaling Theorem 3.1 in \citet{takakura2025optimal} from $[0,1]$ to $[-R,R]$ gives the lower bound $\liminf_{n\to\infty}\risk_n^f\ge16R^4\ve$.
    Combining the two bounds completes the proof.
\end{proof}

\section{Experimental Details}\label{app:experiments}
Here, we give the detailed experimental setup for~\cref{sec:experiments}.
All experiments use $R=1$ and $300{,}000$ trials to estimate the MSE for each condition.

\paragraph{Choice of $C$.}
Here, we describe a principled way of choosing the allocation coefficient $C$.
Let $M_{f,\beta}$ denote the two-phase estimator with preliminary budget fixed at $\beta$, and write $v'_\eta=\odv{v_\eta}{\eta}$.
The normalized MSE has the expansion
\begin{align}
    n^2\Expec{(M_{f,\beta}(D)-f(D))^2}
    =\Delta_f^2v_\varepsilon+\Delta_f^2[-v'_\varepsilon]\beta
    +\frac{K_f}{n^2\beta^4}+o(n^{-2/5}),\label{eq:fixed-budget-risk-allocation-expansion}
\end{align}
when the true means and target are fixed in the interiors of their projection intervals and $\beta=Cn^{-2/5}$.
Here $K_f$ is defined as the second moment of the quadratic preliminary-error.
Minimizing the sum of the second and third terms gives $\beta=(4K_f/(\Delta_f^2[-v'_\varepsilon]))^{1/5}n^{-2/5}$.

The ratios $K_f/\Delta_f^2$ are invariant under rescaling, so set $R=1$ for the following calculation.
From a simple calculation, we obtain
\begin{center}
    \begin{tabular}{@{}clc@{}}
        \toprule
        $f$   & $K_f$                                           & $\sup_D K_f/\Delta_f^2$ \\
        \midrule
        $\mu$ & $40+72\mu^2$                                    & $40$                    \\
        $v$   & $144(v-2\mu^2)^2+1152\mu^2-192v+216$            & $624$                   \\
        $c$   & $320+448(\mu_x^2+\mu_y^2)+240(2\mu_x\mu_y-c)^2$ & $320$                   \\
        \bottomrule
    \end{tabular}
\end{center}
Using the above bounds, we set $C$ as
\begin{align*}
    \begin{aligned}
        C_\mu(\varepsilon) & =\left(\frac{160}{-v'_\varepsilon}\right)^{1/5},\qquad
        C_v(\varepsilon)=\left(\frac{2496}{-v'_\varepsilon}\right)^{1/5}, \quad
        C_c(\varepsilon)   & =\left(\frac{1280}{-v'_\varepsilon}\right)^{1/5}.
    \end{aligned}
\end{align*}
Note that these coefficients depend only on the privacy budget $\varepsilon$.

\subsection{Mean Estimation}\label{app:mean-experiments}
We compare~\cref{alg:mean-estimation} with the hourglass mechanism.
The left panel of~\cref{fig:mean-estimation-performance} fixes $\varepsilon=1$ and $\mu=0$, with
\begin{align*}
    n\in\{30,100,300,1000,3000,10^4,10^5,10^6,10^8\}.
\end{align*}
The right panel fixes $\varepsilon=1$ and $n=10^8$, with
\begin{align*}
    \mu\in\{0,0.25,0.5,0.75,0.9,0.99\}.
\end{align*}

\subsection{Variance and Covariance Estimation}\label{app:variance-covariance-experiments}
For variance and covariance estimation, we consider four specific datasets.
For covariance with $c=0$, the dataset contains $n/4$ copies of $(R, R), (R, -R), (-R, R), (-R, -R)$.
For covariance with $c=1/3$, the dataset contains $n/2$ copies each of $(-1/\sqrt3,-1/\sqrt3)$ and $(1/\sqrt3,1/\sqrt3)$.
For variance with $v=1/3$, the dataset contains $n/2$ copies of $-1/\sqrt3$ and $1/\sqrt3$.
For variance with $v=1/2$, the dataset contains $n/4$ copies of $-1, 1$, $n/2$ copies of zeros.
We use $\varepsilon=1$ in the left panel and $\varepsilon=8$ in the right panel, with
\begin{align*}
    n\in\{20,100,300,1000,3000,10^4,10^5,10^6,10^8\}.
\end{align*}

\section{Additional Results}\label{app:nonzero-moment-results}
\begin{figure}[htbp]
    \centering
    \includegraphics[width=\textwidth]{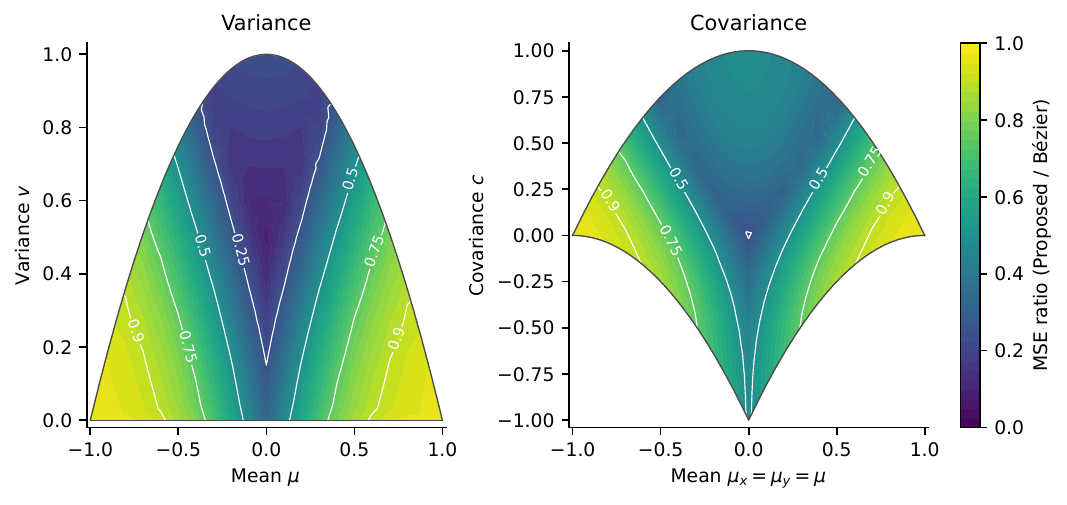}
    \caption{MSE ratios to the B\'ezier mechanisms for variance (left) and covariance (right). Smaller ratios indicate larger improvements.}\label{fig:nonzero-moment-planes}
\end{figure}

Here, we examine how the improvement over the B\'ezier mechanisms varies with dataset statistics.
\Cref{fig:nonzero-moment-planes} plots the empirical MSE ratio of the proposed estimator to the B\'ezier estimator on the $(\mu,v)$ plane for variance and the $(\mu,c)$ plane for covariance, with $\mu_x=\mu_y=\mu$.
In this experiment, we fix $n=10^8$ and $\varepsilon=1$.
We observe that the improvement over the B\'ezier mechanisms is significant for $\mu \simeq 0$
and diminishes as $|\mu|$ approaches $R$.

\end{document}